\documentclass{article}

\PassOptionsToPackage{numbers, compress}{natbib}

 \usepackage[main, final]{neurips_2026}

\usepackage[utf8]{inputenc} 
\usepackage[T1]{fontenc}    
\usepackage{hyperref}       
\usepackage{url}            
\usepackage{booktabs}       
\usepackage{amsfonts}       
\usepackage{nicefrac}       
\usepackage{microtype}      
\usepackage{xcolor}         

\usepackage[pdftex]{graphicx}
\usepackage{subcaption}
\usepackage{multirow}
\usepackage{amsmath}
\usepackage{amssymb}
\usepackage{amsthm}
\usepackage{mathtools}
\usepackage{natbib}
\usepackage{algorithm}
\usepackage{algorithmic}
\usepackage{makecell}

\title{Structure-aware Reinforcement Learning for \\ Protein Directed Evolution}

\author{
Zikun Nie$^{1,2}$, Suyuan Zhao$^{1,2}$, Yizhen Luo$^{1,2}$, Siqi Fan$^{1}$, Zaiqing Nie$^{1,3,}$\thanks{Corresponding author} \\
  $^{1}$Institute of AI Industry Research (AIR), Tsinghua University\\
  $^{2}$Department of Computer Science and Technology, Tsinghua University\\
  $^{3}$PharMolix Inc.\\
  \texttt{\{nzk24,zhaosy23,yz-luo22\}@mails.tsinghua.edu.cn}\\
  \texttt{zaiqing@air.tsinghua.edu.cn} \\
}

\begin{document}

\maketitle

\begin{abstract}
    Protein optimization remains a longstanding goal in life sciences. Existing machine learning–assisted directed evolution (MLDE) methods primarily rely on sequence-only features, overlooking the critical spatial constraints and co-evolutionary interactions encoded in protein structures. However, directly integrating structural information remains challenging due to the scarcity of reliable mutant structures. To address these issues, we propose \textbf{StructEvo}, a novel structure-aware reinforcement learning framework for protein directed evolution. StructEvo employs a delta-structure fusion encoder to approximate mutant structural features via feature differences, enabling effective end-to-end incorporation of spatial knowledge, without predicting coordinate-level structures for individual protein mutants. The vast mutation space is then decomposed into manageable subspaces through a structure-aligned hierarchical action network, while a geometric constraint further stabilizes delta feature learning. Our approach outperforms prior state-of-the-art methods by 9.2\% and 16.3\%, respectively, on two challenging optimization benchmarks, and further identifies an experimentally validated epistasis pattern in GFP, highlighting the importance of structural guidance for effective protein directed evolution.
\end{abstract}

\section{Introduction}
\label{sec:intro}

Proteins are fundamental functional units of biological systems. Compared to \textit{de novo} protein design, optimizing existing proteins through mutations to improve desired functional properties (\textit{i.e.}, \textit{fitness}) offers greater practical value for real-world applications~\cite{soskine2010mutational}, ranging from antibody affinity maturation~\cite{roost1995maturation2, manivel2000maturation} to enzyme catalytic engineering~\cite{anderson2021enzyme,fox2008enzyme}. In practice, the most widely adopted approach for protein optimization is directed evolution~\cite{arnold1998de3, packer2015de2, wang2021directed}, which iteratively modifies proteins by random mutagenesis, expression, screening and selection. This process relies on large-scale \emph{in vivo} expression and \emph{in vitro} screening, making it costly and time-consuming~\cite{yuan2005laboratory}.

\begin{figure}[tbp]
    \centering
    \includegraphics[width=0.9\linewidth]{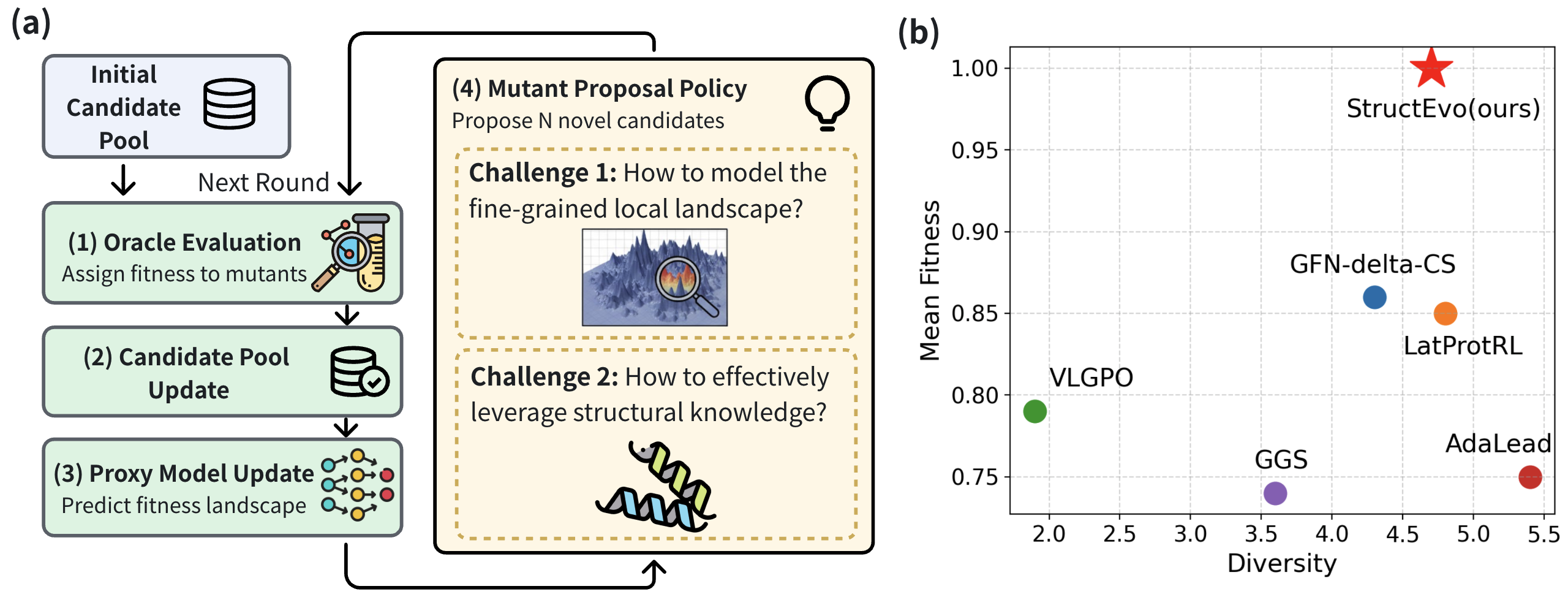}
    \caption{\textbf{(a) Illustration of the active learning pipeline for directed evolution.} Starting from an initial pool, candidates are evaluated by an oracle and then update the pool, after which a proxy is iteratively refined to guide candidate proposals. The figure highlights two key challenges for the proposal policy: how to model the fine-grained local fitness landscape, and how to effectively leverage structural knowledge. \textbf{(b) Performance comparison on GFP-hard benchmark.} StructEvo achieves the highest fitness while maintaining comparable diversity to other high-fitness baselines.}
    \label{fig:task-performance}
    \vskip -0.1in
\end{figure}

Since directed evolution operates on amino acid sequences, existing machine learning-assisted directed evolution (MLDE) approaches~\cite{brookes2019cbas, sinai2020adalead, ren2022pex, lee2024latprotrl, qiu2021clade, qiu2022clade2, wang2024knowrlm} are predominantly sequence-oriented, focusing on modeling the rugged fitness landscape that maps protein sequences to fitness values. Early methods~\cite{qiu2021clade, qiu2022clade2} represent protein sequences by one-hot encodings or hand-crafted features, while recent approaches~\cite{lee2024latprotrl, wang2024knowrlm} leverage protein language models (PLMs) to capture evolutionary priors. However, these sequence-only approaches often overlook the intrinsic spatial context in protein structures, limiting their ability to capture functionally relevant interactions. By explicitly encoding physical and geometric constraints, three-dimensional structures provide rich information about co-evolutionary interaction signals, spatial proximity, and functional motifs, which are essential for protein activity~\cite{hegyi1999relationship}. Motivated by this, we integrate protein structure as a complementary modality to sequence, facilitating more fine-grained geometric guidance for navigating local fitness landscapes.

However, leveraging structural knowledge presents two key challenges: 

(1) \textit{Reliable mutant structures are generally unavailable}. Though a representative experimentally resolved structure may be available for a protein family, mutant structures remain scarce in databases. Meanwhile, existing structure prediction models, such as the AlphaFold~\cite{jumper2021afdb, evans2021afmul, abramson2024af3, cheng2023alphamissense} and Protenix series~\cite{bytedance2025protenix,Zhang2026protenix-v1, Zhang2026protenix-v2}, are computationally costly~\cite{kim2025af3cost, zhu2024scalefold} and often insensitive to point mutations, frequently producing nearly identical structures for similar sequences~\cite{pak2023af2mutation, luppino2025af2mutation2}. Consequently, it remains impractical to directly capture structural variations at the coordinate level, especially in directed evolution settings where even a few selected mutations may induce noticeable backbone deviations. To address this, we model structural changes at the feature level by converting differences in PLM representations, which have been shown to contain rich mutational effects~\cite{luo2024mutaplm}. In this way, we transform the static structure into effective guidance during optimization. 

(2) It remains unclear \textit{how to effectively translate structural signals into mutation decisions}. Jointly optimizing mutation positions and substituted amino acids results in a large action space, making it difficult to learn reliable structural guidance. A key observation is that protein structure is more informative for \textit{where to mutate} than for \textit{mutate to what}, as it provides position-specific spatial context such as residue interactions in three-dimensional space. Motivated by this insight, we decompose mutation decisions into two sequential processes: position selection and amino acid refinement. This design enables structure to guide position decisions, while providing cleaner gradient signals in two low-dimensional subspaces.

Based on these insights, we propose \textbf{StructEvo}, a structure-aware reinforcement learning framework for protein directed evolution. StructEvo introduces a novel paradigm for incorporating structural knowledge into the mutation proposal policy, consisting of three components: (1) a \textit{delta-structure fusion encoder} that approximates mutant structure features via delta representations and integrates sequence and structure information via a cross-attention module; (2) a \textit{structure-aligned hierarchical action network} that decomposes action space and enables structure-guided mutation sampling; and (3) a \textit{geometric constraint} applied on delta representations to improve consistency.

We validate our work through comprehensive experiments on two combinatorial mutation benchmarks and two more challenging full-length benchmarks. Across all benchmarks, StructEvo consistently outperforms strong baselines, and achieves relative fitness gains of up to 9.2\% on AAV-hard and 16.3\% on GFP-hard (Figure~\ref{fig:task-performance}b), demonstrating its effectiveness for protein directed evolution. We further independently discovered a triple-mutation epistasis pattern in GFP validated by existing wet-lab experiments, highlighting the potential of our method for real-world applications. Our code and model are open-sourced at \texttt{https://github.com/Skyyyyyalker/StructEvo}. 

Our contributions are summarized as follows:

\begin{itemize}
\item We propose \textbf{StructEvo}, a structure-aware reinforcement learning framework that introduces a novel paradigm for incorporating structural knowledge into mutation proposal policy.
\item We introduce a \textit{delta-structure fusion encoder}, a \textit{structure-aligned hierarchical action network} and a \textit{geometric constraint} to enable effective structural guidance.
\item We achieve relative fitness gains of 9.2\% and 16.3\% over the prior strongest baselines on two hard benchmarks, and identify a validated triple-mutation pattern in GFP, highlighting the importance of structural guidance for protein directed evolution.
\end{itemize}

\section{Related work}\label{sec:related_work}

\begin{figure*}[tbp]
    \centering
    \includegraphics[width=\linewidth]{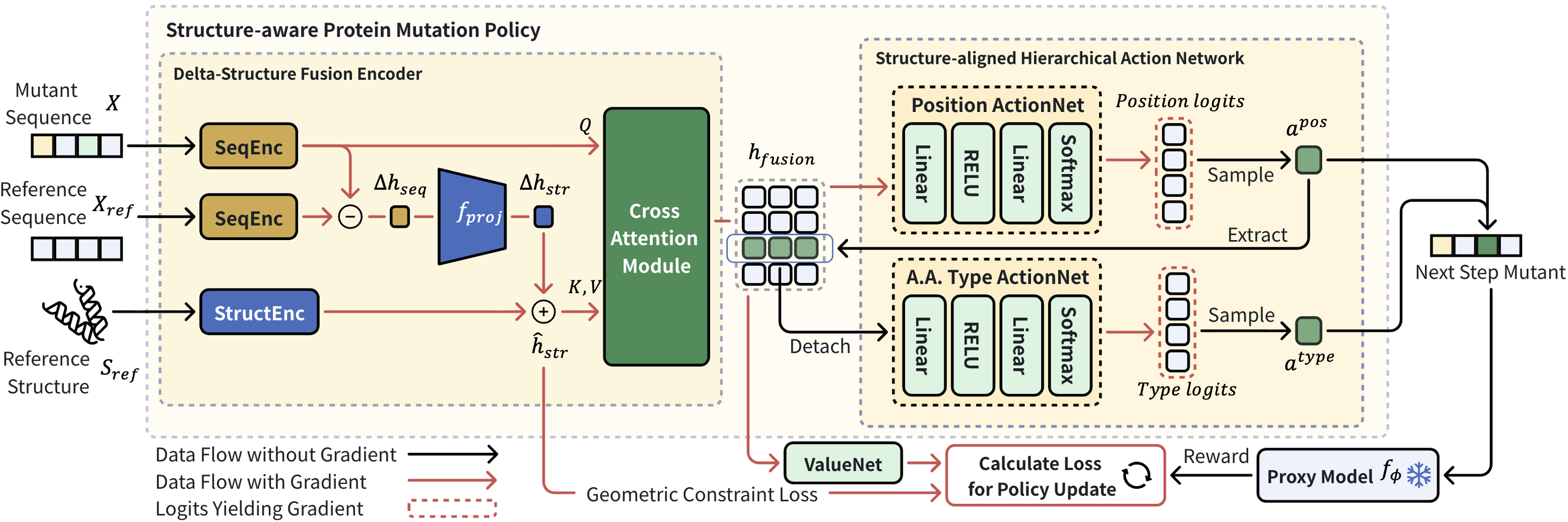}
    \caption{\textbf{Architecture of the StructEvo framework.} A delta-structure fusion encoder approximates mutant structure features via delta representations and integrates sequence and structure information. A structure-aligned hierarchical action network then decomposes action space and enables structure-guided mutation sampling. A geometric constraint loss is calculated to improve robustness. The proxy model remains frozen and provides reward signals during policy updates. A.A.: amino acid.}
    \label{fig:policy}
    \vskip -0.1in
\end{figure*}

\textbf{Machine learning–assisted directed evolution (MLDE).} MLDE aims to iteratively optimize a given starting protein sequence through mutations, where the key challenge is identifying beneficial variants in a rugged local fitness landscape under limited experimental budgets. \emph{Evolutionary search-based methods} simulate natural selection through iterative mutation and recombination~\citep{sinai2020adalead}, with extensions that encourage proximal exploration via distance penalties~\citep{ren2022pex}. \emph{Energy-based methods} learn smooth fitness landscapes as energy models to guide sampling through gradient ascent~\citep{kirjner2024ggs,frey2024dwjs, tran2024latentde}. \emph{Generative model-based methods} explicitly model the mutant distributions using variational autoencoders~\citep{brookes2019cbas,brookes2018dbas} or generative flow networks~\citep{jain2022gflownetsal}, with recent work further introducing controllable generation mechanisms~\citep{bogensperger2025vlgpo, kim2025deltacs}. Existing MLDE methods primarily rely on sequence-only representations, limiting their ability to incorporate structural context during mutation selection.

\textbf{Reinforcement learning (RL) for protein optimization.} Recent studies increasingly formulate protein optimization as a sequential decision-making problem and leverage RL to guide mutation policies~\cite{angermueller2019dynappo}. \citet{wang2023evoplay} generate new sequences through self-play and Monte Carlo tree search, while \citet{wang2024knowrlm} draw physicochemical priors from an amino acid knowledge graph to guide mutation decisions. \citet{lee2024latprotrl} further optimize mutants in a continuous latent space by treating mutations as small perturbations. Unlike prior methods, our approach introduces delta features to capture end-to-end structural variation within a hierarchical RL policy, enabling explicit guidance for proposing high-fitness mutants.

\textbf{Multi-modal protein fitness prediction}. This task is commonly referred to as \textit{fitness landscape modeling}, focusing on predicting functional fitness values for given protein variants~\cite{dallago2021flip, notin2023proteingym}. Existing protein mutation prediction methods focus on learning representations and estimating fitness effects, but do not address the search problem of how to find better mutants. Beyond sequence representations, prior studies incorporate wild-type structure utilizing graph neural networks (GNNs)~\cite{hermosilla2022structure2, chen2023structure} or inverse-folding models~\citep{hsu2022esmif, dauparas2022proteinmpnn, yang2023mifst}. Recent studies further explore sequence-structure fusion through a joint vocabulary~\citep{su2024saprot} or structure quantization module~\citep{li2024prosst}. Additional modalities such as surface topology and multiple sequence alignment (MSA) are also leveraged~\cite{zhang2024s3f,tan2025venusrem}. Compared to static fitness prediction, our work focuses on the more challenging task of proposing beneficial protein mutants in an active learning setting.

\section{Task formulations}
\label{sec:prelim}

\subsection{Directed evolution in an active learning setting}
\label{sec:prelim-active}

In both \textit{in vitro} and \textit{in silico} settings, directed evolution is often conducted in an \emph{active learning} setting~\cite{wang2023evoplay}, where the mutants start from an initial pool and are then optimized iteratively over a total of $R$ rounds. Within each round, the optimization proceeds through four key components (Figure~\ref{fig:task-performance}a):

\textbf{Oracle evaluation.} Let $\mathcal{V}$ denote the vocabulary of the 20 standard amino acids. A protein sequence of length $L$ is defined as $X=(x^1, ..., x^L)\in\mathbb{X}=\mathcal{V}^L$, where each residue $x^i\in\mathcal{V}$. A black-box \textit{oracle} $\mathcal{F}:\mathbb{X}\rightarrow \mathbb{R}$ is a function that maps a protein sequence $X$ to a scalar fitness value $\mathcal{F}(X)$, thereby defining the fitness landscape. In MLDE, the oracle corresponds to costly wet-lab experiments for evaluating candidate sequences. Consequently, the number of oracle queries is strictly limited to $N$ per round, where $N\ll|\mathbb{X}|$.

\textbf{Candidate pool update.} A candidate pool $\mathcal{D}$ is iteratively updated to collect high-fitness mutants. The initial pool $\mathcal{D}_0=\{(X_n, \mathcal{F}(X_n))\}_{n=1}^{|\mathcal{D}_0|}$ contains only a small set of low-fitness candidates. At the beginning of each round $i$, newly proposed mutants are evaluated and incorporated into the pool:
\begin{equation}
    \mathcal{D}_i=\mathcal{D}_{i-1}\cup\{(X_n, \mathcal{F}(X_n))\}_{n=1}^{N}.
\end{equation}

\textbf{Proxy model update.} Due to the limited number of oracle queries, a trainable \textit{proxy} model $f_{\phi}(\cdot)$, parameterized by $\phi$, is used to approximate the landscape defined by the oracle. After updating the candidate pool, the proxy is iteratively fine-tuned on $\mathcal{D}_{i}$ by minimizing the mean squared error loss: 
\begin{equation}
    \mathcal{L}(\phi)=\frac{1}{|\mathcal{D}_i|}\sum_{X\in \mathcal{D}_i}\left(f_\phi(X)-\mathcal{F}(X)\right)^2.
\end{equation}
\textbf{Proposal policy.} Guided by the fine-tuned proxy model $f_\phi$ and starting from the current candidate pool $\mathcal{D}_{i}$, a proposal policy generates a batch of $N$ novel mutants at each round. These mutants are then evaluated by the oracle and subsequently update the candidate pool for the next round.

The main objective of directed evolution is to \textit{identify a batch of $K$ high-fitness protein mutants under a fixed budget of $R$ rounds}. A summary of all symbols is provided in Appendix~\ref{app:symbol} for reference.

\subsection{Reinforcement learning-based directed evolution}
\label{sec:prelim-rlde}

We formulate protein optimization as a reinforcement learning (RL) problem. By modeling sequential mutation trajectories as a Markov Decision Process (MDP), RL enables learning a policy that sequentially selects mutations to achieve high fitness.

\textbf{State.} At step $t$, the state $s_t=(x_t^1, ..., x_t^L)$ is a protein mutant with $L$ residues, where $x_t^i \in \mathcal{V}$.

\textbf{Action.} An action $a_t=(a_t^\text{pos}, a_t^\text{type})$ at step $t$ represents a single-point mutation applied to the current state $s_t$. The action $a_t$ can be factorized into two parts: the mutation position $a_t^\text{pos}$ and the type of substituted amino acid $a_t^\text{type}$. In our task, the transition function $P(s_{t+1}\mid s_t, a_t)$ is deterministic and the state is updated by:
\[
x_{t+1}^i =
\begin{cases}
a_t^\text{type}, & i = a_t^\text{pos}, \\
x_t^i, & \text{otherwise}.
\end{cases}
\]
\textbf{Reward.} Reward signals are provided by the proxy model. In our work, we adopt a sparse reward setting, where the reward $r_t=r(s_t, a_t)$ is assigned only at trajectory termination. A trajectory terminates either when the policy finds a mutant better than the starting one, or when a predefined maximum number of steps $m_\text{step}$ is reached. Specifically, for a trajectory ending at step $T$, the reward is $r_T=f_{\phi}(s_{T+1})$, while intermediate rewards are set to $r_t=0$ for $t<T$. We emphasize that rewards are generated solely by the proxy model, and the oracle is used only for final evaluation.

\textbf{Training Objective.} We aim to learn a policy $\pi(a_t\mid s_t)$ that maximizes the expected return over trajectories. Under the sparse reward setting, the training objective is formulated as:
\begin{equation}
\max_\pi \; \mathbb{E}_{\tau \sim \pi} \left[ \sum_{t=0}^{T} \gamma^t r_t \right]
= \max_\pi \; \mathbb{E}_{\tau \sim \pi} \left[ \gamma^T f_{\phi}(s_{T+1}) \right],
\end{equation}
where $\tau = (s_0, a_0, \dots, s_T, a_T)$ denotes a mutation trajectory terminating at step $T$, and $\gamma$ is the discount factor. We optimize the policy using Proximal Policy Optimization (PPO)~\cite{schulman2017ppo}, which has demonstrated effectiveness in prior studies~\cite{lee2024latprotrl, wang2024knowrlm}. The collected candidates are then ranked by proxy-predicted fitness and evaluated by the oracle for subsequent refinement in the next round.

Within the reinforcement learning framework, our work primarily focuses on improving the key component---the \textit{mutation proposal policy}, as described in the following section.

\section{Methodology}
\label{sec:method}

The main goal of our work is to incorporate structural knowledge into the protein mutation policy that maps states to actions. In this section, we present the proposed StructEvo framework, highlighting three main components: (1) a \textit{delta-structure fusion encoder} that fuses sequential and structural features for a given mutant (Sec.~\ref{sec:method-delta}), (2) a \textit{structure-aligned hierarchical action network} that receives the fused feature and samples high-fitness mutations (Sec.~\ref{sec:method-hierarchical}), and (3) a \textit{geometric constraint} between delta features to improve robustness and consistency (Sec.~\ref{sec:method-geoloss}).

\subsection{Delta-structure fusion encoder}
\label{sec:method-delta}

The delta-structure fusion encoder integrates both sequence and structural knowledge to produce a fine-grained representation of protein mutants. As illustrated in Figure~\ref{fig:policy}, it consists of a protein sequence encoder $\mathcal{E}_\text{seq}$, a protein structure encoder $\mathcal{E}_\text{str}$, a geometric projector $f_\text{proj}$ and a cross-attention module. We adopt ESM-2~\cite{lin2022esm2}, a protein language model (PLM) pretrained on large-scale evolutionary sequence data, to initialize the sequence encoder and capture mutant sequence features. For structure encoding, we employ ESM-IF~\cite{hsu2022esmif} due to its simplicity and efficiency, as it captures both topology and geometric constraints of protein structures. Notably, the StructEvo framework is model-agnostic and easily extensible to different protein encoders.

\textbf{Formulation of delta sequence feature.} Since reliable mutant structures are generally unavailable, we adopt a \textit{reference structure} $S^\text{ref}$ with the highest reliability as a structural scaffold for the corresponding mutant family (see Appendix~\ref{app:structure-process}), with its sequence denoted as the \textit{reference sequence} $X^\text{ref}$. Prior work~\cite{luo2024mutaplm} has shown that differences between PLM representations capture rich information of mutational effect. Therefore, we introduce the delta sequence feature ${\Delta}h_\text{seq}$ as:
\begin{equation}
  \Delta h_{\text{seq}} = h_\text{seq} - h_\text{seq}^\text{ref} = \mathcal{E}_\text{seq}(X)-\mathcal{E}_\text{seq}(X^\text{ref}).  
\end{equation}
$X,X^\text{ref}$ denote the sequences of the current and the reference proteins, and $h_\text{seq},h_\text{seq}^\text{ref}$ are the corresponding sequence representations, where $d_\text{seq}$ denotes its dimensionality.

\textbf{Projection into structural manifold.} We speculate that the mutational knowledge contained in the delta sequence feature reflects corresponding structural changes. Accordingly, we introduce a projector $f_\text{proj}: \mathbb{R}^{L \times d_\text{seq}} \rightarrow \mathbb{R}^{L \times d_\text{str}}$ to map the delta sequence feature from the sequence space to the structural manifold, yielding the delta structure feature $\Delta h_\text{str}$. Both the sequence and structural representations are aligned to residue-level embeddings of the protein backbone, enabling us to get the approximated mutant structural features $\hat{h}_\text{str}$ by adding the projected delta features to the reference structural features. These procedures are formulated as:
\begin{equation}
\Delta h_\text{str} = f_\text{proj}({\Delta}h_\text{seq}),\quad
\hat{h}_\text{str} = h^\text{ref}_\text{str} + {\Delta}h_{\text{str}} = \mathcal{E}_\text{str}(S^\text{ref}) + {\Delta}h_{\text{str}}.
\end{equation}
Here, $S^\text{ref}\in \mathbb{R}^{L\times3\times3}$ is the backbone coordinates of the reference structure, where each residue is represented by the three-dimensional positions of its N, C$\alpha$, and C atoms. To ensure sequence-structure consistency, we further impose a geometric constraint between delta features, as described below in Sec.~\ref{sec:method-geoloss}.

\textbf{Structure-conditioned feature fusion.} To integrate sequence and structure information, we employ a cross-attention module where the mutant sequence feature $h_\text{seq}$ serves as the query, and the approximated structure feature $\hat{h}_\text{str}$ serves as the key and value. The asymmetric design allows sequence features to selectively query and attend to structural context. The resulting fusion feature $h_\text{fusion}$ encodes both sequential and structural information of the mutant, serving as input to the downstream hierarchical action network for mutation sampling.

\subsection{Structure-aligned hierarchical action network}
\label{sec:method-hierarchical}

In RL formulation, the Markov decision process decomposes the vast mutation space into a sequence of single-step decision problems. Each action corresponds to a single-point mutation $a = (a^\text{pos}, a^\text{type})$, where $a^\text{pos}\in\{1, ..., L\}$ denotes the mutation position and $a^\text{type} \in \mathcal{V}$ denotes the substituted amino acid type. Here, $L$ is the protein length and $\mathcal{V}$ is the standard amino acid vocabulary. For the four-site benchmarks, mutation positions are restricted to the four designated sites. As depicted in Figure~\ref{fig:policy}, the decision process is decomposed into two hierarchical steps:

\textbf{Mutation position sampling.} Since mutation position is more tightly coupled with the structural context than amino acid type, prioritizing position selection $a^\text{pos}$ enables the policy to better exploit structure-derived insights. A \textit{position action network} $\pi_{\theta_1}$ receives the fusion feature $h_\text{fusion}$ as input and outputs a categorical distribution over mutation position subspace, from which the position action $a^\text{pos}\sim \pi_{\theta_1}(\cdot \mid h_\text{fusion})$ is sampled. 

\textbf{Substituted amino acid sampling.} Once the position $a^\text{pos}$ is determined, the corresponding local representation at the specific position is then \textit{extracted} and \textit{detached} from the fusion feature. An \textit{amino acid type action network} $\pi_{\theta_2}$ then operates on the detached context to sample the substituted amino acid $a^\text{type}\sim \pi_{\theta_2}(\cdot \mid h_\text{fusion}, a^\text{pos})$. Here, the \textit{detach} operation blocks gradients backpropagating into the shared fusion encoder and the position network. This prevents gradients from the amino acid type network from directly updating the shared fusion representation, allowing the position network to receive a cleaner and more focused learning signal (see Appendix~\ref{sec:app-detach-proof}).

\subsection{Geometric constraint for delta features}
\label{sec:method-geoloss}

\textbf{Geometric constraint.} In the delta-structure fusion encoder, the delta sequence feature is projected to obtain a corresponding delta structure feature. We aim to maintain consistency between structural variations and sequence perturbations, based on the intuition that small mutational effects captured by PLM should not lead to excessively large changes in structural features, and vice versa. To encourage this, we introduce a per-residue geometric constraint between the two delta features, formulated as:
\begin{equation}
    \mathcal{L}_\text{geo}=\mathbb{E}_{\Delta h_\text{str}, \Delta h_\text{seq}}\left[
    \frac{1}{L}\sum^{L}_{i=1}{\left(
    \frac{\|\Delta h^{i}_\text{str}\|}{\sqrt{d_\text{str}}}
    - c * \frac{\|\Delta h^{i}_\text{seq}\|}{\sqrt{d_\text{seq}}}
    \right)^2}~\right],
\end{equation}
where $c$ is the scaling factor, and $\|\cdot\|$ denotes the $\ell_2$-norm. From another perspective, the geometric constraint is inspired by the notion of \textit{Lipschitz continuity}~\cite{asadi2018lipschitz}, as it encourages the magnitude of structural feature variations to remain proportional to that of sequence perturbations, thus enabling a more stable and faithful mapping (see Appendix~\ref{sec:app-lipschitz-proof}).

\textbf{Overall training objective.} The overall RL training objective consists of two parts: 
\begin{equation}
    \mathcal{L}_\text{total} = \underbrace{
        \mathcal{L}_\text{policy}
        + \lambda_\text{entropy} \cdot \mathcal{L}_\text{entropy}
        + \lambda_\text{value} \cdot \mathcal{L}_\text{value}
    }_\text{standard PPO loss}
    + \underbrace{\lambda_\text{geo} \cdot \mathcal{L}_\text{geo}}_\text{geometric constraint}.
\end{equation}
This objective jointly optimizes the mutation policy while enforcing consistency between delta sequence and structure features. The policy loss $\mathcal{L}_\text{policy}$, entropy loss $\mathcal{L}_\text{entropy}$, and value loss $ \mathcal{L}_\text{value}$ are defined in the same way as in the standard PPO loss function~\cite{schulman2017ppo}; see Appendix~\ref{sec:app-loss} for details. Algorithm~\ref{alg:policy} summarizes the overall mutation proposal pipeline.
\section{Experiments}\label{sec:exp}

In this section, we demonstrate that StructEvo is effective on directed evolution tasks across two combinatorial benchmarks (Sec.~\ref{sec:exp-4site}) and two more challenging full-length mutation benchmarks (Sec.~\ref{sec:exp-full-len}). We further provide an in-depth analysis of our model (Sec.~\ref{sec:exp-ablation}) and a case study that reveals high-fitness mutation patterns in green fluorescent protein (Sec.~\ref{sec:exp-case}).

\subsection{Optimization on combinatorial benchmarks}
\label{sec:exp-4site}

\begin{table*}[t]
\caption{\textbf{Results on GB1 and PhoQ benchmarks}. We report average results and std across 5 independent runs. $^*$: results are derived from \citet{wang2024knowrlm}. n.a.: not reported in the original paper.}
\label{tab:result-4site}
\centering
\resizebox{\textwidth}{!}{
\begin{tabular}{lcccccc}
    \toprule
    & \multicolumn{3}{c}{GB1} & 
    \multicolumn{3}{c}{PhoQ} \\
    \cmidrule(r){2-4} \cmidrule(r){5-7}
    Method & 
    Max$\uparrow$ & Mean$\uparrow$ & NDCG$\uparrow$ & 
    Max$\uparrow$ & Mean$\uparrow$ & NDCG$\uparrow$ \\
    \midrule
    CMA-ES      & 0.59$\pm$0.17 & 0.13$\pm$0.04 & 0.75$\pm$0.02 
                & 0.29$\pm$0.14 & 0.05$\pm$0.01 & 0.69$\pm$0.02 \\
    BO          & 0.67$\pm$0.10 & 0.17$\pm$0.02 & 0.77$\pm$0.01 
                & 0.29$\pm$0.06 & 0.05$\pm$0.01 & 0.70$\pm$0.01 \\
    AdaLead     & 0.64$\pm$0.13 & 0.30$\pm$0.09 & 0.74$\pm$0.02 
                & 0.27$\pm$0.11 & 0.14$\pm$0.02 & 0.67$\pm$0.02 \\
    PEX         & 0.66$\pm$0.03 & 0.36$\pm$0.02 & 0.76$\pm$0.01 
                & 0.40$\pm$0.03 & 0.10$\pm$0.02 & 0.71$\pm$0.02 \\
    MLDE$^*$      & 0.68$\pm$n.a. & 0.20$\pm$n.a. & 0.79$\pm$n.a. 
                  & 0.36$\pm$n.a. & 0.10$\pm$n.a. & 0.79$\pm$n.a. \\
    ftMLDE (EVmutation)$^*$   
                  & 0.94$\pm$n.a. & 0.41$\pm$n.a. & 0.83$\pm$n.a. 
                  & 0.44$\pm$n.a. & 0.11$\pm$n.a. & 0.80$\pm$n.a. \\
    ftMLDE (Transformer)$^*$  
                  & 0.93$\pm$n.a. & 0.42$\pm$n.a. & 0.81$\pm$n.a. 
                  & 0.42$\pm$n.a. & 0.11$\pm$n.a. & \underline{0.82$\pm$n.a.} \\
    EvoPlay$^*$   & 0.84$\pm$n.a. & 0.46$\pm$n.a. & 0.85$\pm$n.a. 
                  & 0.47$\pm$n.a. & 0.14$\pm$n.a. & 0.80$\pm$n.a. \\
    CLADE$^*$     & 0.84$\pm$n.a. & 0.46$\pm$n.a. & 0.86$\pm$n.a. 
                  & 0.47$\pm$n.a. & 0.09$\pm$n.a. & 0.78$\pm$n.a. \\
    CLADE2.0$^*$ & 0.94$\pm$n.a. & 0.49$\pm$n.a. & \underline{0.88$\pm$n.a.}
                  & 0.40$\pm$n.a. & 0.15$\pm$n.a. & 0.81$\pm$n.a. \\
    KnowRLM$^*$ &
    \underline{0.97$\pm$0.06} & \underline{0.56$\pm$0.02} & \underline{0.88$\pm$0.02} & 
    \textbf{0.66$\pm$n.a.}    & \underline{0.16$\pm$n.a.} & \underline{0.82$\pm$n.a.} \\
    
    \midrule
    \textbf{StructEvo (ours)} &
    \textbf{0.99$\pm$0.01} & \textbf{0.59$\pm$0.01} & \textbf{0.93$\pm$0.01} & 
    \underline{0.64$\pm$0.07} & \textbf{0.19$\pm$0.02} & \textbf{0.86$\pm$0.01} \\
    \bottomrule
\end{tabular}}
\end{table*}

\textbf{Datasets.} We first evaluate StructEvo on two widely recognized four-site combinatorial datasets, GB1 and PhoQ. The GB1 dataset~\citep{wu2019mlde} consists of 149,361 variants at four epistatic sites (V39, D40, G41 and V54) of the GB1 domain of streptococcal protein G, with fitness reflecting binding to the Fc region of human IgG. Only 2.4\% variants have higher fitness than the wild-type protein, highlighting the sparsity of the GB1 fitness landscape. The PhoQ dataset~\citep{podgornaia2015phoq} contains 140,517 variants of the PhoQ histidine kinase mutated at four key interfacial residues (A284, V285, S288 and T289), with fitness reflecting both kinase and phosphatase responsiveness of PhoP/PhoQ regulatory system on signal transduction. Compared to GB1, the PhoQ fitness landscape is even sparser---98.7\% mutants with fitness below 0.1 and 63.7\% with zero, making it a particularly challenging task. All fitness values are min-max normalized to [0, 1]. Details of benchmarks are provided in Appendix~\ref{app:4site-benchmarks}.

\textbf{Baselines.} We compare our method with various strong baselines, including (1) probability-based methods, BO~\citep{wilson2017bo} and CMA-ES~\citep{hansen2003cmaes}; (2) evolutionary search-based methods, AdaLead~\citep{sinai2020adalead}, PEX~\citep{ren2022pex}, MLDE~\citep{wu2019mlde} and two variants of ftMLDE~\citep{wittmann2021ftmlde}; (3) cluster-based methods, CLADE~\citep{qiu2021clade} and its advanced version CLADE2.0~\citep{qiu2022clade2}, and (4) recent RL-based methods, EvoPlay~\citep{wang2023evoplay} and KnowRLM~\citep{wang2024knowrlm}. Implementation details can be found in Appendix~\ref{app:baselines-4site}. 

\textbf{Metrics.} We report the \emph{maximum} and \emph{mean} fitness over a combined set of candidates, including $N$ initial mutants, $N*R$ proposed mutants across $R$ rounds, and the top $N$ mutants ranked by the final proxy. Here, $R$ denotes the total optimization rounds and $N$ is the number of proposed candidates per round. Given that the ground truth fitness is fully available in combinatorial tasks, we further evaluate the ranking quality of proxy by normalized discounted cumulative gain (NDCG). Formal definitions of metrics are provided in Appendix~\ref{app:metric-4site}.

\textbf{Experimental settings.} Following prior work~\citep{wang2024knowrlm}, we set $R=3$ and $N=96$. The starting $N$ mutants are sampled by CLADE~\citep{qiu2021clade}. The oracle is the ground-truth dataset, with rare missing instances assigned a value of 0. All experiments are repeated over 5 independent runs. 

\textbf{Results and analysis.} We present performance comparisons at round 3 in Table~\ref{tab:result-4site}, with more detailed results of each round in Appendix~\ref{app:full-results-4site}. We observe that: (1) StructEvo achieves state-of-the-art performance on GB1, outperforming the strongest RL-based baseline (KnowRLM) by relative gains of 5.7\% on NDCG, demonstrating its strong capability in navigating fine-grained fitness landscapes and guiding mutant proposals. (2) Though the reference structure of PhoQ is predicted by AlphaFold2~\cite{jumper2021afdb} (with an average pLDDT of 82.94, see Appendix~\ref{app:structure-process}) rather than experimentally resolved, StructEvo still remains competitive with KnowRLM on PhoQ benchmark, with relative gains of 4.9\% on NDCG and 18.8\% on mean fitness. This observation shows the robustness of our framework, while also indicating the importance of accurate structural information for landscape exploration. (3) Probabilistic and evolutionary methods perform poorly under extremely scarce data. The key strength of PEX, proximal preference, becomes less significant in the constrained four-site space. Cluster-based methods leverage prior knowledge to reduce aimless exploration, while RL-based methods further improve performance by explicitly modeling the sequential decision-making process. Hence, we emphasize the significance of combining structural priors for effective protein optimization in data-scarce settings.

\begin{table}[t]
\caption{\textbf{Performance on AAV and GFP benchmarks}. The results include average and standard deviation across 5 independent runs. $^*$: results are derived from~\citet{kirjner2024ggs}. $^\dag$: results are derived from~\citet{lee2024latprotrl}. -: not reported in the original paper.}
\label{tab:result-full-len}
\centering
\resizebox{\linewidth}{!}{
\begin{tabular}{lcccccccc}
    \toprule
     & \multicolumn{4}{c}{\emph{AAV medium}}
     & \multicolumn{4}{c}{\emph{AAV hard}} \\
    \cmidrule(r){2-5} \cmidrule(r){6-9}
    Method & 
    Mean$\uparrow$ & Max$\uparrow$ & Diversity & Novelty &
    Mean$\uparrow$ & Max$\uparrow$ & Diversity & Novelty \\
    \midrule
    CMA-ES & 
    0.05$\pm$0.00 & 0.40$\pm$0.01 & 20.9$\pm$0.5 & 17.2$\pm$0.5 & 
    0.05$\pm$0.00 & 0.32$\pm$0.03 & 20.4$\pm$0.7 & 18.6$\pm$0.5 \\
    BO & 
    0.64$\pm$0.03 & 0.72$\pm$0.04 & 8.1$\pm$0.2 & 9.4$\pm$0.9 & 
    0.62$\pm$0.03 & 0.70$\pm$0.03 & 8.6$\pm$0.2 & 10.3$\pm$0.3 \\
    PEX & 
    0.65$\pm$0.01 & 0.74$\pm$0.00 & 5.5$\pm$0.5 & 5.3$\pm$0.3 & 
    0.63$\pm$0.02 & 0.75$\pm$0.01 & 5.7$\pm$0.5 & 6.2$\pm$0.4 \\
    GGS$^*$ &
    0.51$\pm$0.01 & - & 4.0$\pm$0.2 & 5.4$\pm$0.5 &
    0.60$\pm$0.02 & - & 4.5$\pm$0.5 & 7.0$\pm$0.0 \\
    AdaLead & 
    0.74$\pm$0.04 & 0.80$\pm$0.05 & 5.2$\pm$0.9 & 7.6$\pm$0.4 & 
    \underline{0.76}$\pm$0.03 & \underline{0.81$\pm$0.03} & 4.0$\pm$0.4 & 8.9$\pm$0.6 \\
    VLGPO &
    0.69$\pm$0.01 & 0.76$\pm$0.01 & 3.1$\pm$0.2 & 5.7$\pm$0.1 &
    0.71$\pm$0.01 & 0.78$\pm$0.01 & 3.3$\pm$0.2 & 6.9$\pm$0.1 \\
    LatProtRL$^\dag$ & 
    0.71$\pm$0.02 & - & 5.4$\pm$0.8 & 6.0$\pm$0.2 & 
    0.66$\pm$0.01 & - & 6.0$\pm$0.8 & 7.0$\pm$0.6 \\
    GFN-$\delta$-CS &
    \underline{0.77$\pm$0.02} & \underline{0.81$\pm$0.02} & 4.4$\pm$1.1 & 5.8$\pm$1.5 &
    \underline{0.76$\pm$0.03} & 0.80$\pm$0.03 & 4.0$\pm$0.7 & 6.7$\pm$1.1 \\
    \midrule
    \textbf{StructEvo} & 
    \textbf{0.82$\pm$0.01} & \textbf{0.86$\pm$0.01} & 4.5$\pm$0.7 & 8.6$\pm$0.2 & 
    \textbf{0.83$\pm$0.03} & \textbf{0.87$\pm$0.04} & 4.5$\pm$0.3 & 8.4$\pm$0.5 \\
    \midrule
     & \multicolumn{4}{c}{\emph{GFP medium}}
     & \multicolumn{4}{c}{\emph{GFP hard}} \\
    \cmidrule(r){2-5} \cmidrule(r){6-9}
    Method & 
    Mean$\uparrow$ & Max$\uparrow$ & Diversity & Novelty &
    Mean$\uparrow$ & Max$\uparrow$ & Diversity & Novelty \\
    \midrule
    CMA-ES & 
    -0.04$\pm$0.01 & 0.58$\pm$0.00 & 172.7$\pm$2.1 & 154.3$\pm$4.9 & 
    -0.03$\pm$0.00 & 0.16$\pm$0.03 & 173.8$\pm$2.0 & 218.6$\pm$7.6 \\
    BO & 
    0.30$\pm$0.07 & 0.42$\pm$0.07 & 26.4$\pm$5.6 & 37.7$\pm$12.1 & 
    0.26$\pm$0.07 & 0.39$\pm$0.04 & 36.0$\pm$2.7 & 71.7$\pm$10.1 \\
    PEX & 
    0.64$\pm$0.04 & 0.84$\pm$0.04 & 8.3$\pm$2.6 & 6.8$\pm$1.9 & 
    0.46$\pm$0.05 & 0.62$\pm$0.05 & 7.0$\pm$1.2 & 10.3$\pm$1.7 \\
    GGS$^*$ &
    0.76$\pm$0.01 & - & 3.7$\pm$0.2 & 5.0$\pm$0.1 &
    0.74$\pm$0.00 & - & 3.6$\pm$0.0 & 6.9$\pm$0.1 \\
    AdaLead & 
    0.93$\pm$0.01 & 0.99$\pm$0.02 & 5.2$\pm$0.9 & 9.5$\pm$1.2 & 
    0.75$\pm$0.04 & 0.81$\pm$0.04 & 5.4$\pm$0.6 & 13.5$\pm$1.1 \\
    VLGPO &
    0.93$\pm$0.03 & 1.02$\pm$0.01 & 2.8$\pm$0.2 & 5.3$\pm$0.0 &
    0.79$\pm$0.02 & \underline{0.95$\pm$0.00} & 1.9$\pm$0.0 & 6.2$\pm$0.0\\
    LatProtRL$^\dag$ & 
    0.93$\pm$0.00 & - & 4.6$\pm$0.4 & 5.5$\pm$0.1 & 
    0.85$\pm$0.01 & - & 4.8$\pm$0.3 & 7.0$\pm$0.4 \\
    GFN-$\delta$-CS &
    \underline{1.06$\pm$0.04} & \underline{1.09$\pm$0.03} & 7.0$\pm$0.4 & 9.5$\pm$1.9 &
    \underline{0.86$\pm$0.06} & 0.90$\pm$0.05 & 4.3$\pm$1.6 & 8.2$\pm$1.5 \\
    \midrule
    \textbf{StructEvo} & 
    \textbf{1.11$\pm$0.02} & \textbf{1.17$\pm$0.02} & 6.3$\pm$0.8 & 11.9$\pm$0.8 & 
    \textbf{1.00$\pm$0.04} & \textbf{1.05$\pm$0.05} & 4.9$\pm$0.3 & 11.4$\pm$0.6 \\
    \bottomrule
\end{tabular}
}
\vskip -0.1in
\end{table}

\begin{figure*}[tbp]
  \begin{center}
    \centerline{\includegraphics[width=\columnwidth]{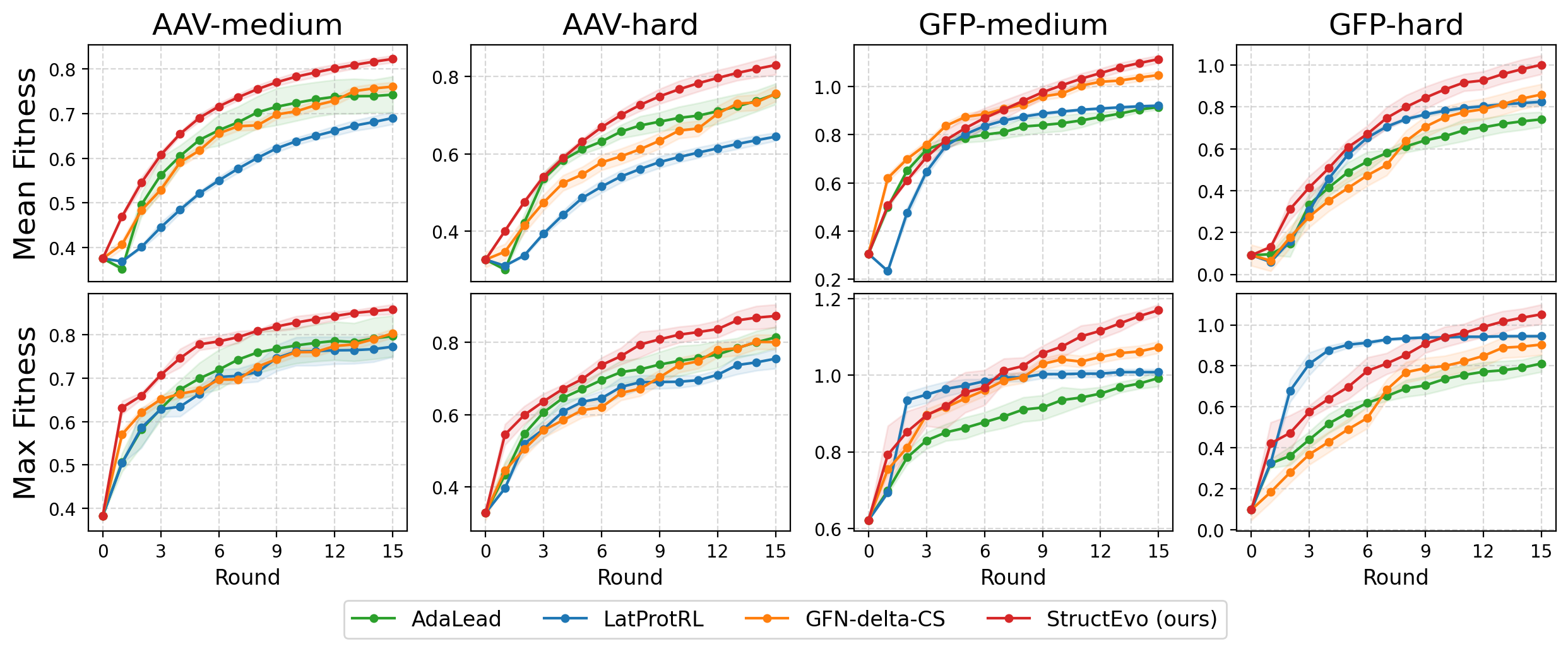}}
    \caption{\textbf{Visualization of mean (top row) and maximum (bottom row) fitness across rounds.} The curves indicate the average results, and the shaded regions indicate the standard deviation.}
    \label{fig:aav-gfp-best3}
  \end{center}
  \vskip -0.3in
\end{figure*}

\subsection{Optimization on full-length benchmarks}
\label{sec:exp-full-len}

\textbf{Datasets.} We further evaluate StructEvo on two full-length protein mutation benchmarks, \textit{GFP} and \textit{AAV}. Compared to combinatorial tasks, these are more challenging as mutations span the full sequence, leading to an exponentially larger search space. The GFP dataset~\citep{sarkisyan2016gfp} measures log-fluorescence intensity of 56,806 \emph{Aequorea victoria} green fluorescent protein variants, with a mutation space size of $|\mathbb{X}|=20^{237}$. The AAV dataset~\citep{bryant2021aav} focuses on engineering a 28-amino acid segment of the VP1 capsid protein of adeno-associated virus, comprising 44,156 variants from a space of $|\mathbb{X}|=20^{28}$. Following~\citet{kirjner2024ggs}, we partition both benchmarks into \textit{medium} and \textit{hard} regimes. The initial fitness distribution under the \textit{hard} setting is more distant from the high-fitness range than the \textit{medium} setting, leading to increased optimization difficulty. See more details in Appendix~\ref{app:full-length-benchmarks}.

\textbf{Baselines.} In addition to the baselines introduced for combinatorial benchmarks, we further compare StructEvo with (1) the energy model-based methods, GGS~\citep{kirjner2024ggs}; (2) latent-space optimization methods, LatProtRL~\citep{lee2024latprotrl} and VLGPO~\citep{bogensperger2025vlgpo}; and (3) the GFlowNet-based method, GFN-$\delta$-CS~\citep{kim2025deltacs}. Implementation details are provided in Appendix~\ref{app:baselines-full-length}. Since MLDEs~\cite{wu2019mlde, wittmann2021ftmlde} and CLADEs~\cite{qiu2021clade, qiu2022clade2} need to enumerate all possible mutants to construct a global scoring vector, they may not be suitable for full-length benchmarks, where the exponentially large mutation space size ($20^L$) makes such enumeration infeasible.

\textbf{Metrics.} We report the \emph{mean} and \emph{maximum} fitness of the top $N$ sequences proposed over $R$ rounds. Fitness values are evaluated by the oracle and min-max normalized using the ground truth dataset. Notably, normalized fitness may exceed 1.0, as the oracle can assign higher fitness to previously unseen mutants than the maximum observed in the ground-truth dataset. We additionally choose distance-based metrics, \textit{diversity} and \textit{novelty}, as described in Appendix~\ref{app:metric-full-length}.

\textbf{Experimental settings.} Following prior work~\cite{kirjner2024ggs}, we set the total rounds to $R=15$, the number of candidates proposed per round, i.e., the evaluation budget, to $N=128$. The oracle models are derived from ~\citet{kirjner2024ggs}. Each method is run with 5 independent seeds.

\textbf{Results and analysis.} Table~\ref{tab:result-full-len} and Figure~\ref{fig:aav-gfp-best3} show comparisons between StructEvo and baselines on full-length benchmarks. We observe that: (1) StructEvo consistently achieves state-of-the-art fitness across all four scenarios, surpassing the strongest baseline by relative gains of 9.2\% and 16.3\% in mean fitness on the AAV and GFP \textit{hard} settings, respectively. Furthermore, the larger improvements on \textit{hard} settings compared to \textit{medium} ones highlight the increasing importance of structural guidance when high-fitness data is severely limited. (2) While LatProtRL converges more quickly during the early rounds on GFP, StructEvo continues to improve across rounds and ultimately surpasses it. This indicates that structural information could provide sustained guidance for exploring more fine-grained high-fitness regions in the long-horizon optimization. Improving early-round efficiency remains an important direction for future work. (3) We note that higher diversity and novelty \textit{do not} simply imply better performance, as pointed out in previous work~\citep{lee2024latprotrl}. For instance, CMA-ES attains the highest diversity and novelty but yields dysfunctional mutants with near-zero fitness. In contrast, StructEvo achieves the highest fitness while maintaining comparable or higher diversity and novelty than other high-fitness methods, reflecting a more effective trade-off between exploration and exploitation.

\subsection{Ablation studies}
\label{sec:exp-ablation}

\begin{figure}[t]
\centering
\begin{minipage}{0.65\linewidth}
\centering
\captionof{table}{\textbf{Results of ablation studies.} $X,X^\text{ref}$: mutant and reference sequence, $S^\text{ref}$: reference structure. w/o: without.}
\label{tab:ablation}
\resizebox{\linewidth}{!}{
\begin{tabular}{ccp{0.4cm}p{0.4cm}p{0.4cm}ccc}
    \toprule
    \multirow{2}{*}{} & \multirow{2}{*}{Settings} & \multicolumn{3}{c}{\emph{Inputs}} & \multirow{2}{*}{Mean$\uparrow$} & \multirow{2}{*}{Max$\uparrow$} & \multirow{2}{*}{Time(s)$\downarrow$} \\
    \cmidrule(r){3-5}
    & & $X$ & $X^\text{ref}$ & $S^\text{ref}$ &  &  &  \\
    \midrule
    \multirow{5}{0.6cm}{\centering \emph{AAV}\\ \emph{hard}}
    &StructEvo  & \checkmark  & \checkmark & \checkmark & \textbf{0.83} & \textbf{0.87}  &  0.337 \\
    &(1)        & \checkmark & &\checkmark & 0.80 & 0.84 & 0.323 \\
    &(2)        & \checkmark & \checkmark & & 0.75 & 0.79 & 0.297 \\
    &(3)        & \checkmark & & & 0.75 & 0.81 & 0.303 \\
    &(4)  & \multicolumn{3}{c}{w/o hierarchy} & 0.77 & 0.81 & 0.583 \\
    \midrule
    \multirow{5}{0.6cm}{\centering \emph{GFP}\\ \emph{hard}}
    &StructEvo  & \checkmark & \checkmark & \checkmark & \textbf{1.00} & \textbf{1.05}  &  0.825 \\
    &(1)        & \checkmark &  & \checkmark & 0.97 & 1.01 & 0.819 \\
    &(2)        & \checkmark & \checkmark &  & 0.83 & 0.89 & 0.753 \\
    &(3)        & \checkmark &  &  & 0.84 & 0.93 & 0.697 \\
    &(4)  & \multicolumn{3}{c}{w/o hierarchy} & 0.75 & 0.82 & 3.515 \\
    \bottomrule
\end{tabular}
}

\end{minipage}
\hfill
\begin{minipage}{0.34\linewidth}
\centering
\includegraphics[width=\linewidth]{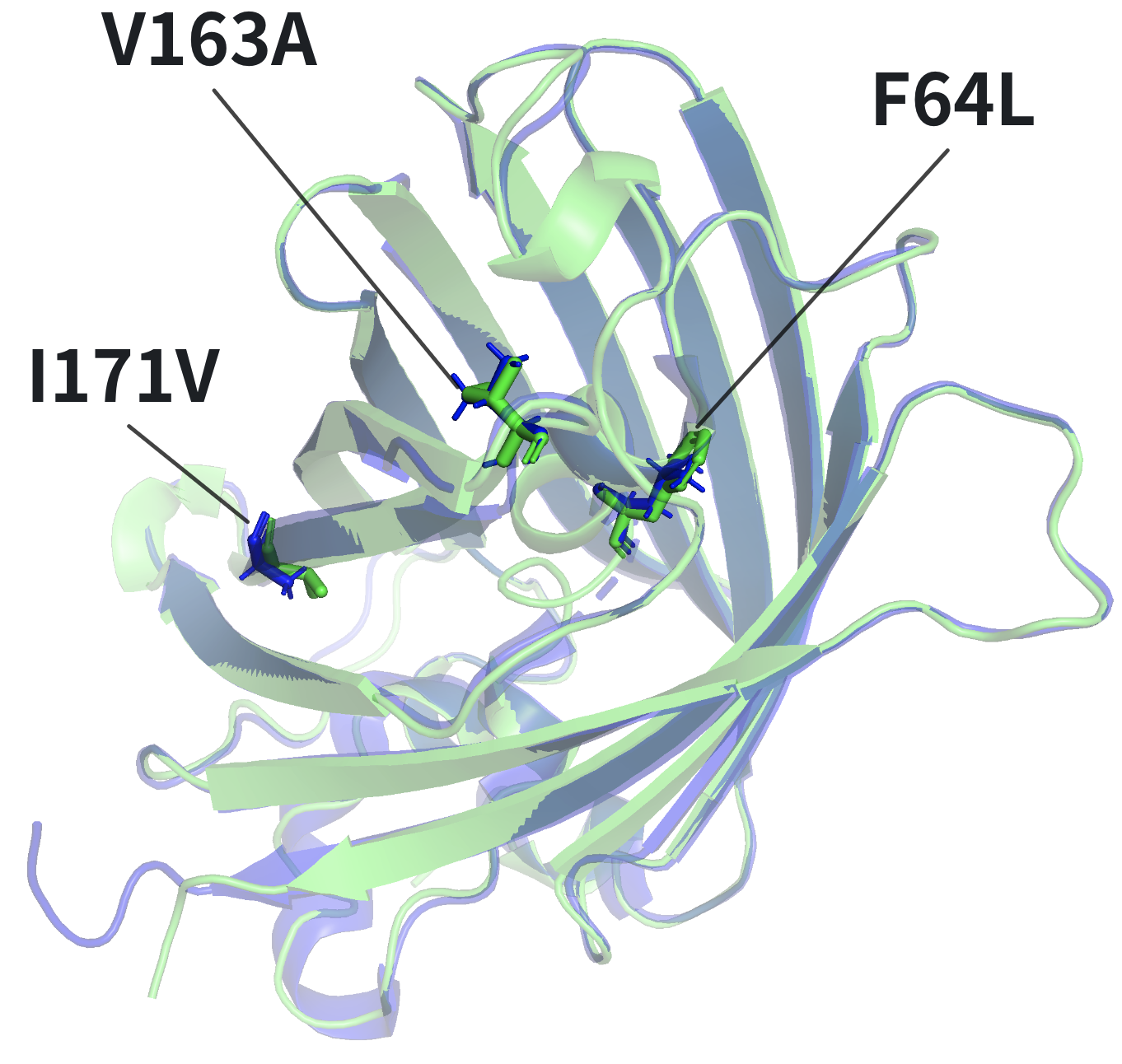}
\caption{\textbf{A triple-mutation mutant in GFP.} The reference (green) and the mutant (blue) structure are rendered by PyMOL.}
\label{fig:case}
\end{minipage}
\vskip -0.15in
\end{figure}

We examine the impact of each component in StructEvo by: (1) removing the delta module by directly fusing mutant sequence with the static reference structure via cross-attention, (2) removing structural inputs and fusing two sequence features, (3) relying solely on mutant sequence, and (4) replacing the hierarchical action network with a flat network that jointly samples both actions. We also report the runtime cost per candidate for each setting. As shown in Table~\ref{tab:ablation} and Appendix~\ref{app:full-ablation}, removing delta module degrades performance, while removing structural information leads to a larger drop (9.6\% mean fitness on AAV-hard and 17.0\% on GFP-hard). Substituting structure with reference sequence yields no improvements. Moreover, sampling in the joint space significantly degrades fitness and incurs much longer runtime, with both drawbacks amplified as protein length increases. These findings validate each component in our framework. We refer readers to more case studies and additional in-depth analyses of structure prediction models, model contribution isolation, proxy robustness, and geometric constraint sensitivity in Appendix~\ref{app:additional-indepth}.

\subsection{Case study: a triple-mutation epistasis in GFP}
\label{sec:exp-case}

In GFP, we identify a triple-mutation pattern F64L/V163A/I171V (PDB: 8DPD; Figure~\ref{fig:case}). From 3,840 proposed mutants over 15 rounds, 271 mutants with this triple mutation achieve substantially higher fitness (average 0.91) than 1,834 single (0.07) and 427 double mutants (0.23), revealing a pronounced \textit{epistasis effect}. Interestingly, this pattern has been validated and explained by wet-lab experiments~\cite{delagrave1995gfpval1,pedelacq2006gfpval2}: residue 64 is near the chromophore, while 163 and 171 are located in the proximal region of $\beta$-barrel; all mutations result in smaller side-chains, potentially facilitating tighter packing through synergistic effects. This promotes energy dissipation via photon emission and ultimately boosts fluorescence intensity. We further find a higher cosine similarity of this true mutant structure feature to the approximated feature (0.85) than to the reference (0.73), supporting the validity of delta module. Notably, none of the baselines recover this pattern, highlighting the advantage of structure in capturing long-range co-evolutionary signals that are typically missed by sequence-only approaches.


\section{Conclusions}
\label{sec:conclusion}

In this work, we propose StructEvo, a structure-aware reinforcement learning framework for protein directed evolution. By approximating mutant structure features via delta representations and decomposing the action space via a hierarchical action network, StructEvo introduces a new paradigm for incorporating structural knowledge into mutation proposal. Comprehensive experiments on both combinatorial and full-length benchmarks show consistent performance improvements over strong baselines, demonstrating the value of dynamic structural context for efficient protein optimization. Several limitations remain: (1) extending to protein complexes rather than monomers; (2) expanding single-objective fitness to multi-objective properties; and (3) integrating feedback from wet-lab experiments, which we leave for future work. While StructEvo shows potential in real-world applications, we emphasize safety concerns that \textit{any protein variants generated by our method should undergo rigorous experimental validation and careful human inspection before practical use.}

\section*{Acknowledgments}

This research is supported by the Innovative Drug Research and Development National Science and Technology Major Project (No.2025ZD1803101), the Wuxi Research Institute of Applied Technologies, Tsinghua University (Grant 20242001120), and PharMolix Inc.

\bibliography{section/refs}

\newpage
\appendix

\section*{Appendix}

\section{Details of StructEvo}
\subsection{Symbol definitions}
\label{app:symbol}

\begin{table}[h]
\centering
\caption{Definition of symbols used in our work.}
\label{tab:app-symbol}
\resizebox{\linewidth}{!}{
\begin{tabular}{ccl}
    \toprule
    Category & Symbol & Definition \\
    \midrule
    \multirow{12}{2cm}{\centering \textit{Protein} \\ \textit{Representation}}
    & $L$ & the protein length \\
    & $X$ & the protein sequence \\
    & $S$ & the protein structure \\
    & $\mathbb{X}$ & the mutation space \\
    & $\mathcal{V}$ & the amino acid vocabulary \\
    & $x_i$ & the amino acid identity of residue $i$ \\
    & $a_i,b_i,c_i$ & the three-dimensional coordinates of residue $i$ \\
    & $\mathcal{E}_{seq},\mathcal{E}_{str}(\cdot)$ & the sequence and structure encoder \\
    & $h_\text{seq},h_\text{str},h_\text{fusion}$ & the sequence feature, structure feature, and the fusion feature \\
    & $\Delta h_\text{seq}, \Delta h_\text{str}$ & the delta sequence and structure feature \\
    & $\hat{h}_\text{str}$ & the approximated mutant structure feature \\
    & $d_\text{seq},d_\text{str}$ & the hidden size of sequence and structure features \\
    \midrule
    \multirow{6}{2cm}{\centering \textit{Active}\\ \textit{Learning}}
    & $R$ & the number of total rounds \\
    & $N$ & the number of proposed candidates per round \\
    & $\mathcal{F}(\cdot)$ & the oracle function \\
    & $f_\phi(\cdot)$ & the proxy model parameterized by $\phi$ \\
    & $\mathcal{D}$ & the mutant candidate pool \\
    & $\mathcal{D}_0$ & the initial mutant candidate pool \\
    \midrule
    \multirow{9}{2cm}{\centering \textit{Reinforcement}\\ \textit{Learning} \\ \textit{and}\\ \textit{Policy}}    
    & $\mathbb{A}$ & the action space \\
    & $s_t$ & the state at step $t$ \\
    & $a_t$ & the action at step $t$ \\
    & $a_t^{pos}, a_t^{type}$ & the mutation position and substituted amino acid type at step $t$ \\
    & $r_t$ & the reward calculated by the proxy at step $t$ \\
    & $\gamma$ & the discount rate \\
    & $\pi_{\theta_1}, \pi_{\theta_2}$ & the parameterized position and amino acid type action networks \\
    & $\mathcal{T}, \tau$ & the collected trajectory \\
    & $m_{step}$ & the maximum number of steps in a trajectory \\
    \bottomrule
\end{tabular}
}
\end{table}

\subsection{Algorithm for StructEvo framework pipeline}
\label{app:algorithm}

We provide a full description of the StructEvo pipeline in Algorithm~\ref{alg:policy}. The outer loop is the active learning process, where selected mutant candidates are evaluated by the oracle and then iteratively update the candidate pool, followed by proxy model fine-tuning. The inner loop shows the reinforcement learning procedure within each round, including how the mutation trajectories are collected, how rewards are assigned, and how the policy network is updated. 

In the RL loop, the trajectory terminates once an improved variant is identified. This design choice simplifies the implementation and provides better control. More importantly, it ensures that each round focuses on relatively short-distance mutations, keeping the proxy model within its current trust region to provide reliable prediction guidance. Longer-distance mutations are achieved through multiple rounds.

The construction of the initial pool $\mathcal{D}_0$ differs slightly between the 4-site tasks and full-length tasks. For 4-site tasks, CLADE~\cite{qiu2021clade}, an unsupervised hierarchical clustering strategy, is used to sample $K$ candidates from the combinatorial space as the initial pool. The initial sequence is then randomly selected from it. For the full-length tasks, the initial pools are already provided (i.e., initial training sets), and the initial sequence is randomly selected from the top $K$ candidates ranked by oracle fitness. For both task types, we set $K$ to be the same as the oracle budget $N$.

\begin{algorithm}[htbp]
  \caption{The StructEvo pipeline}
  \label{alg:policy}
  \begin{algorithmic}[1]
    \STATE {\bfseries Input:} Total rounds $R$, number of proposed candidates per round $N$, black-box oracle $\mathcal{F}$, starting candidates set $\mathcal{D}_0=\{X_n, \mathcal{F}(X_n)\}_{n=1}^{|\mathcal{D}_0|}$, reference structure $S^\text{ref}$ and sequence $X^\text{ref}$, sequence encoder $\mathcal{E}_\text{seq}$, structure encoder $\mathcal{E}_\text{str}$, trajectory max length $m_\text{step}$, total max step $m_\text{total}$
    
    \STATE Initialize current round $i\leftarrow 0$
    \WHILE{$i < R$}
        \STATE {\color{blue} \textit{// active learning loop begin}}
        \STATE Fine-tune the proxy model $f_\phi$ with validated candidates $\mathcal{D}_i$
        \STATE Initialize candidate buffer $\mathcal{D} \leftarrow \emptyset$
        \STATE Initialize trajectory buffer $\mathcal{T}\leftarrow \emptyset$
        \STATE Initialize total steps $t_\text{total} \leftarrow 0$
        \WHILE{$t_\text{total} < m_\text{total}$}
            \STATE {\color{blue} \textit{// RL loop begin}}
            \STATE Sample an initial sequence $X_0 \in \mathcal{D}_{i}$
            \STATE Initialize state $s_0 \leftarrow X_0$, reward threshold $r_\text{init} \leftarrow f_\phi(s_0)$
            \STATE step $t \leftarrow 0$, trajectory $\tau\leftarrow \emptyset$, ${\text{FLAG}_\text{stop}} \leftarrow \texttt{False}$
            \WHILE{not $\text{FLAG}_\text{stop}$}
                \STATE {\color{blue} \textit{// delta-structure fusion encoder}}
                \STATE Delta sequence feature $\Delta h_{\text{seq},t} \leftarrow \mathcal{E}_\text{seq}(s_t)-\mathcal{E}_\text{seq}(X^\text{ref})$
                \STATE Delta structure feature $\Delta h_{\text{str}, t} \leftarrow f_\text{proj}({\Delta}h_{\text{seq}, t})$
                \STATE Approximated mutant structure feature $\hat{h}_{\text{str},t} \leftarrow \mathcal{E}_\text{str}(S^\text{ref}) + {\Delta}h_{\text{str}, t}$
                \STATE Calculate geometric constraint loss $\mathcal{L}_\text{geo}(\Delta h_{\text{str}, t}, \Delta h_{\text{seq}, t})$
                \STATE Fusion feature $h_{\text{fusion}, t}\leftarrow\texttt{CrossAttention}(\mathcal{E}_\text{seq}(s_t), \hat{h}_{\text{str},t}, \hat{h}_{\text{str},t})$
                \STATE {\color{blue} \textit{// hierarchical action network}}
                \STATE Sample mutation position $a_t^\text{pos} \sim \pi_{\theta_1}(\cdot \mid h_{\text{fusion}, t})$
                \STATE Sample substituted amino acid type $a_t^\text{type} \sim \pi_{\theta_2}(\cdot \mid h_{\text{fusion}, t}, a_t^\text{pos})$
                \STATE Apply mutation to obtain next state $s_{t+1} \leftarrow \texttt{transition}(s_t, (a_t^\text{pos}, a_t^\text{type}))$
                \STATE Assign reward $r_t\leftarrow f_\phi(s_{t+1})$
                \IF{$r_t>r_\text{init}$ \OR $t\geq m_\text{step}$}
                \STATE $\text{FLAG}_\text{stop}\leftarrow\texttt{True}$
                \ENDIF
                \IF{not $\text{FLAG}_\text{stop}$}
                    \STATE $r_t\leftarrow 0$ \quad \textit{// sparse reward for intermediate steps}
                \ENDIF
                \STATE $\tau\leftarrow\tau\cup\{s_t, a_t, r_t\}$
                \STATE $t \leftarrow t+1$
                \STATE $t_\text{total} \leftarrow t_\text{total}+1$
            \ENDWHILE
            \STATE Update $\mathcal{T} \leftarrow \mathcal{T} \cup \tau$
            \STATE Update $\mathcal{D}\leftarrow \mathcal{D} \cup \{s_t\}$
        \ENDWHILE
        \STATE Update policy by $\mathcal{L}_\text{PPO} + \lambda_\text{geo}\cdot\mathcal{L}_\text{geo}$ with collected trajectories $\mathcal{T}$
        \STATE Select $N$ novel candidates ranked by proxy $\{X_n\}^N_{n=1} \leftarrow \mathtt{TopN}(\mathcal{D}, f_\phi(\mathcal{D}))$
        \STATE Evaluate $\{X_n\}^N_{n=1}$ by oracle $\mathcal{F}$
        \STATE Update candidate pool $\mathcal{D}_{i+1}\leftarrow \mathcal{D}_i \cup \{X_n, \mathcal{F}(X_n)\}^N_{n=1}$
        \STATE $i \leftarrow i+1$
    \ENDWHILE
    \STATE return proposed candidates $\mathcal{D}_R$
  \end{algorithmic}
\end{algorithm}

\subsection{The PPO loss function}
\label{sec:app-loss}

The policy, entropy, and value losses follow the standard PPO formulation~\citep{schulman2017ppo}. Specifically,

$$
\mathcal{L}_{policy}=-\mathbb{E}_t\left[\min\left(r_t(\theta)A_t,\operatorname{clip}(r_t(\theta),1-\epsilon,1+\epsilon)A_t\right)\right],
$$

$$
\mathcal{L}_{value}=\mathbb{E}_t\left[\left(V_\Phi(s_t)-R_t\right)^2\right],
$$

$$
\mathcal{L}_{entropy}=-\mathbb{E}_t\left[H(\pi_\theta(\cdot|s_t))\right].
$$

Here, $r_t(\theta)=\frac{\pi_\theta(a_t|s_t)}{\pi_{\theta_{\mathrm{old}}}(a_t|s_t)}$ denotes the policy probability ratio, $A_t$ is the estimated advantage, $V_\Phi(s_t)$ is the value function, $R_t$ is the return target, and $H(\pi_\theta(\cdot|s_t))$ represents the policy entropy. Other notations follow the original PPO paper.

\subsection{The detach operator}
\label{sec:app-detach-proof}

As introduced in Sec.~\ref{sec:method-hierarchical}, in the proposed structure-aligned hierarchical action network, the localized feature is extracted and detached from the fusion feature. We next show how the \textit{detach} operation blocks gradient flow and leads to a cleaner learning signal. For simplicity, we use $h$ below to denote $h_\text{fusion}$ in the main text.

As introduced in ~\citet{schulman2017ppo}, the training objective of PPO consists of three components:
\begin{equation}
    L_t^{\mathrm{CLIP+VF+S}}(\theta)
= \hat{\mathbb{E}}_t \Big[
L_t^{\mathrm{CLIP}}(\theta)
- c_1 L_t^{\mathrm{VF}}(\theta)
+ c_2 S\big[\pi_\theta\big](s_t)
\Big],
\end{equation}

where the main objective $L_t^{\mathrm{CLIP}}(\theta)$ is:
\begin{equation}
    L_t^{\mathrm{CLIP}}(\theta)
= \hat{\mathbb{E}}_t \Big[
\min\big(
r_t(\theta)\,\hat{A}_t,\;
\operatorname{clip}\big(r_t(\theta),\,1-\epsilon,\,1+\epsilon\big)\,\hat{A}_t
\big)
\Big],
\end{equation}

with the probability ratio $r_t(\theta)$ defined as:
\begin{equation}
    r_t(\theta)=\frac{\pi_\theta(a_t \mid s_t)}{\pi_{\theta_{\mathrm{old}}}(a_t \mid s_t)}.
\end{equation}

In our hierarchical action network, the mutation action $a_t$ is decomposed into $a_t=(a_t^\text{pos}, a_t^\text{type})$, which is sampled sequentially through two action networks $\pi_{\theta_1}$ and $\pi_{\theta_2}$. Therefore, the probability of sampled action $a_t$ at given state $s_t$ corresponds to:
\begin{equation}
    \pi_\theta(a_t \mid s_t) = p(a_t^\text{pos}, a_t^\text{type}) = \pi_{\theta_1}(a_t^\text{pos} \mid h) \cdot \pi_{\theta_2}(a_t^\text{type} \mid h, a_t^\text{pos}).
\end{equation}

During policy optimization, gradients of the PPO objective are computed. We focus on the joint log-likelihood component $\nabla_{\theta_1,\theta_2}\log p(a^\text{pos}, a^\text{type})$ associated with the action network. Under the hierarchical regime, it factorizes as:
\begin{align}
&\nabla_{\theta_1,\theta_2}\log p(a^\text{pos}, a^\text{type}) \\
&=\nabla_{\theta_1,\theta_2}\log\left[ \pi_{\theta_1}(a^\text{pos} \mid h) \cdot \pi_{\theta_2}(a^\text{type} \mid h, a^\text{pos})\right]\\
&=\nabla_{\theta_1,\theta_2}\left[ \log \pi_{\theta_1}(a^\text{pos} \mid h) + \log \pi_{\theta_2}(a^\text{type} \mid h, a^\text{pos})\right]\\
&=\nabla_{\theta_1} \log \pi_{\theta_1}(a^\text{pos} \mid h) + \nabla_{\theta_2}\log \pi_{\theta_2}(a^\text{type} \mid h, a^\text{pos}),
\end{align}

where the sampling operation blocks the gradients of the amino acid decision back-propagating to the position action network. The gradients from $\pi_{\theta_2}$ cannot influence the preceding feature fusing module, since the detaching operation enforces $\nabla_{h}~\pi_{\theta_2}( a^\text{type} \mid h, a^\text{pos})=\textbf{0}$. As a result, the position network $\pi_{\theta_1}$ is optimized to capture the overall contribution of selecting mutation sites, yielding a cleaner and more stable learning signal for position selection. This decomposition enables the structure-enhanced features to more directly and effectively guide the identification of functional mutation positions, thereby reducing learning complexity and improving data efficiency under limited experimental feedback. 

\subsection{The geometric constraint}
\label{sec:app-lipschitz-proof}

In Sec.~\ref{sec:method-geoloss}, we introduce a per-residue geometric constraint loss between the delta sequence feature $\Delta h_\text{seq} \in \mathbb{R}^{L \times d_\text{seq}}$ and delta structure feature $\Delta h_\text{str} \in \mathbb{R}^{L \times d_\text{str}}$, formulated as:
\begin{equation}
    \mathcal{L}_\text{geo}(\Delta h_\text{str}, \Delta h_\text{seq})
    =\frac{1}{L}\sum^{L}_{i=1}{\left(
    \frac{\|\Delta h^{i}_\text{str}\|}{\sqrt{d_\text{str}}}
    - c * \frac{\|\Delta h^{i}_\text{seq}\|}{\sqrt{d_\text{seq}}}
    \right)^2},
\end{equation}
where $c$ is the scaling factor. Here, we provide an intuitive explanation of how the proposed geometric constraint encourages reference-centered norm consistency, analogous to a Lipschitz-type upper bound. We refer readers to \citet{asadi2018lipschitz} for a more rigorous analysis on how Lipschitz continuity can promote smoother mappings and more stable learning signals in RL framework.

For simplicity, we denote the delta feature matrices as:
\begin{equation}
    \mathbf{x}=\Delta h_\text{seq}=\{x_{ij}\}_{i={1, ..., L}}^{j=1, ..., d_\text{seq}}, \quad
    \mathbf{y}=\Delta h_\text{str}=\{y_{ij}\}_{i={1, ..., L}}^{j=1, ..., d_\text{str}}.
\end{equation}
The corresponding $\ell_2$ norms are defined as:
\begin{equation}
    \|\mathbf{x}\|=\sqrt{\sum_{i=1}^{L}\sum_{j=1}^{d_\text{seq}}x_{ij}^2}, \quad
    \|\mathbf{x}_i\|=\sqrt{\sum_{j=1}^{d_\text{seq}}x_{ij}^2},  \quad
    \|\mathbf{y}\|=\sqrt{\sum_{i=1}^{L}\sum_{j=1}^{d_\text{str}}y_{ij}^2},  \quad
    \|\mathbf{y}_i\|=\sqrt{\sum_{j=1}^{d_\text{str}}y_{ij}^2}. 
\end{equation}
Under this notation, the geometric constraint loss is formulated as:
\begin{equation}
\mathcal{L}_\text{geo} = \frac{1}{L} \sum_{i=1}^L (\frac{\|\mathbf{y}_i\|}{\sqrt{d_\text{str}}} - c \cdot \frac{\|\mathbf{x}_i\|}{\sqrt{d_\text{seq}}})^2.
\end{equation}
We consider the idealized global minimum where $\mathcal{L}_\text{geo} = 0$. This implies:
\begin{equation}
    \|\mathbf{y}_i\| = \left(c \cdot \frac{\sqrt{d_\text{str}}}{\sqrt{d_\text{seq}}}\right) \cdot \|\mathbf{x}_i\| \quad \text{for all} \quad i \in \{1, \dots, L\}.
\end{equation}
We denote $c_0= c \cdot \frac{\sqrt{d_\text{str}}}{\sqrt{d_\text{seq}}}$. Squaring both sides, we obtain:
\begin{equation}
    \sum_{j=1}^{d_\text{str}}y_{ij}^2=\|\mathbf{y}_i\|^2 = c_0^2 \cdot \|\mathbf{x}_i\|^2 = c_0^2 \cdot \sum_{j=1}^{d_\text{seq}}x_{ij}^2
    \quad \text{for all} \quad i \in \{1, \dots, L\}.
\end{equation}
Summing over $i$ from $1$ to $L$, we obtain:
\begin{equation}
    \sum_{i=1}^{L}\sum_{j=1}^{d_\text{str}}y_{ij}^2 = \sum_{i=1}^{L}c_0^2 \cdot \|\mathbf{x}_i\|^2 = c_0^2 \cdot \sum_{i=1}^{L}\sum_{j=1}^{d_\text{seq}}x_{ij}^2.
\end{equation}
Thus,
\begin{equation}
    \|\mathbf{y}\|^2 = c_0^2 \cdot \|\mathbf{x}\|^2.
\end{equation}
Therefore,
\begin{equation}\label{eq:2}
    \|\mathbf{y}\| = c_0 \cdot \|\mathbf{x}\|.
\end{equation}
We define the projection mapping from delta sequence manifold $\mathbf{x}=\Delta h_\text{seq}$ to delta structure manifold $\mathbf{y}=\Delta h_\text{str}$ as $\mathcal{M}: \mathbf{x} \mapsto \mathbf{y}$. By definition, the mapping $\mathcal{M}$ is \textit{$K$-Lipschitz continuous} if any input $\mathbf{x}$ satisfies:
\begin{equation}\label{eq:1}
    \| \mathcal{M}(\mathbf{x}) - \mathcal{M}(\mathbf{x}^\text{ref}) \| \le K \cdot \| \mathbf{x} - \mathbf{x}^\text{ref} \|.
\end{equation}
In the context of our delta representation, any delta feature is defined as the difference $\Delta h_\text{seq}=h_\text{seq}-h_\text{seq}^\text{ref}$. Accordingly, the delta feature of the reference protein is $\mathbf{x}^\text{ref}=\Delta h_\text{seq}^\text{ref}=h_\text{seq}^\text{ref}-h_\text{seq}^\text{ref}=\mathbf{0}$. This simplifies Eq.~\ref{eq:1} to:
\begin{equation}\label{eq:3}
     \| \mathbf{y} \| = \| \mathcal{M}(\mathbf{x}) \| \le K \cdot \| \mathbf{x} \|.
\end{equation}
Comparing Eq.~\ref{eq:2} and Eq.~\ref{eq:3}, we meet the upper bound of the Lipschitz continuity with the Lipschitz constant: 
\begin{equation}
    K=c_0 = c \cdot \frac{\sqrt{d_\text{str}}}{\sqrt{d_\text{seq}}}.
\end{equation}
Thus, the geometric constraint loss $\mathcal{L}_\text{geo}$ enables a similar mechanism with Lipschitz continuity, encouraging consistent sensitivity in the projection mapping and leading to more stable learning signals for RL training process. We also conduct experiments to investigate the sensitivity of this geometric constraint, as detailed in Appendix~\ref{app:geo_coef}.

\section{Protein fitness landscape modeling}

Protein fitness landscape modeling aims to learn a fine-grained mapping from proteins to fitness values. 

\textbf{Sequence-based models}~\citep{rao2021msatransformer, meier2021esm1v, lin2022esm2, brandes2022proteinbert, notin2022tranception} formulate protein residues as discrete tokens analogous to natural language, and are pre-trained on large-scale sequence databases via masked language modeling (MLM;~\citealp{devlin2019bert}), enabling zero-shot fitness prediction. 

\textbf{Structure-based models}~\citep{hermosilla2022structure2,chen2023structure} typically leverage Graph Neural Network (GNN;~\citealp{wu2020gnn}) to explicitly encode three-dimensional protein structures as nodes and edges. Inverse-folding models~\citep{hsu2022esmif, dauparas2022proteinmpnn, yang2023mifst} leverage structural representations to predict protein sequence. 

More recently, \textbf{hybrid models} have incorporated both sequence and structure to obtain more informative protein representations. \citet{su2024saprot} introduced a concatenated structure-aware vocabulary, while \citet{li2024prosst} proposed a structure quantization module with a disentangled attention mechanism for modality fusion.

\section{Experimental details}

\subsection{Combinatorial mutation benchmarks}
\label{app:4site-benchmarks}

The GB1 and PhoQ datasets are summarized at Table~\ref{tab:app-gb1-dataset}. Fitness of GB1 dataset measures the binding affinity of human protein GB1 domain to the antibody IgG-Fc, while fitness of PhoQ refers to kinase and phosphatase responsiveness of PhoP/PhoQ regulatory system. The key advantage of these datasets is that they are sufficiently large to almost exhaustively cover the theoretical four-site combinatorial space ($20^4=160,000$), thereby serving as reliable assessments for MLDE methods.

\begin{table}[h]
    \caption{\textbf{Description of GB1 and PhoQ benchmarks.} The last three columns report the proportion of mutants whose normalized fitness is lower than a given threshold, highlighting the sparsity and challenge of these benchmarks. aa: amino acids.}
    \label{tab:app-gb1-dataset}
    \centering
    \begin{tabular}{cccccccc}
        \toprule
        Name    & \makecell{UniProt\\ID} & \makecell{Protein\\Length} & \makecell{Dataset\\Size} & \makecell{Mutation\\Sites} & \textless 0.3 & \textless 0.1 & = 0 \\
        \midrule
        \textbf{GB1}      & \texttt{P19909} & 56aa & 149,361 & \makecell{V39,D40,\\G41,V54}     & 99.30\% & 97.30\% & 19.74\% \\
        \midrule
        \textbf{PhoQ}     & \texttt{P23837} & 486aa & 140,517 & \makecell{A284,V285,\\S288,T289} & 99.96\% & 98.65\% & 63.70\% \\
        \bottomrule
    \end{tabular}
\end{table}

\subsection{Full-length mutation benchmarks}
\label{app:full-length-benchmarks}

\begin{figure}[htbp]
    \centering
    \includegraphics[width=0.7\linewidth]{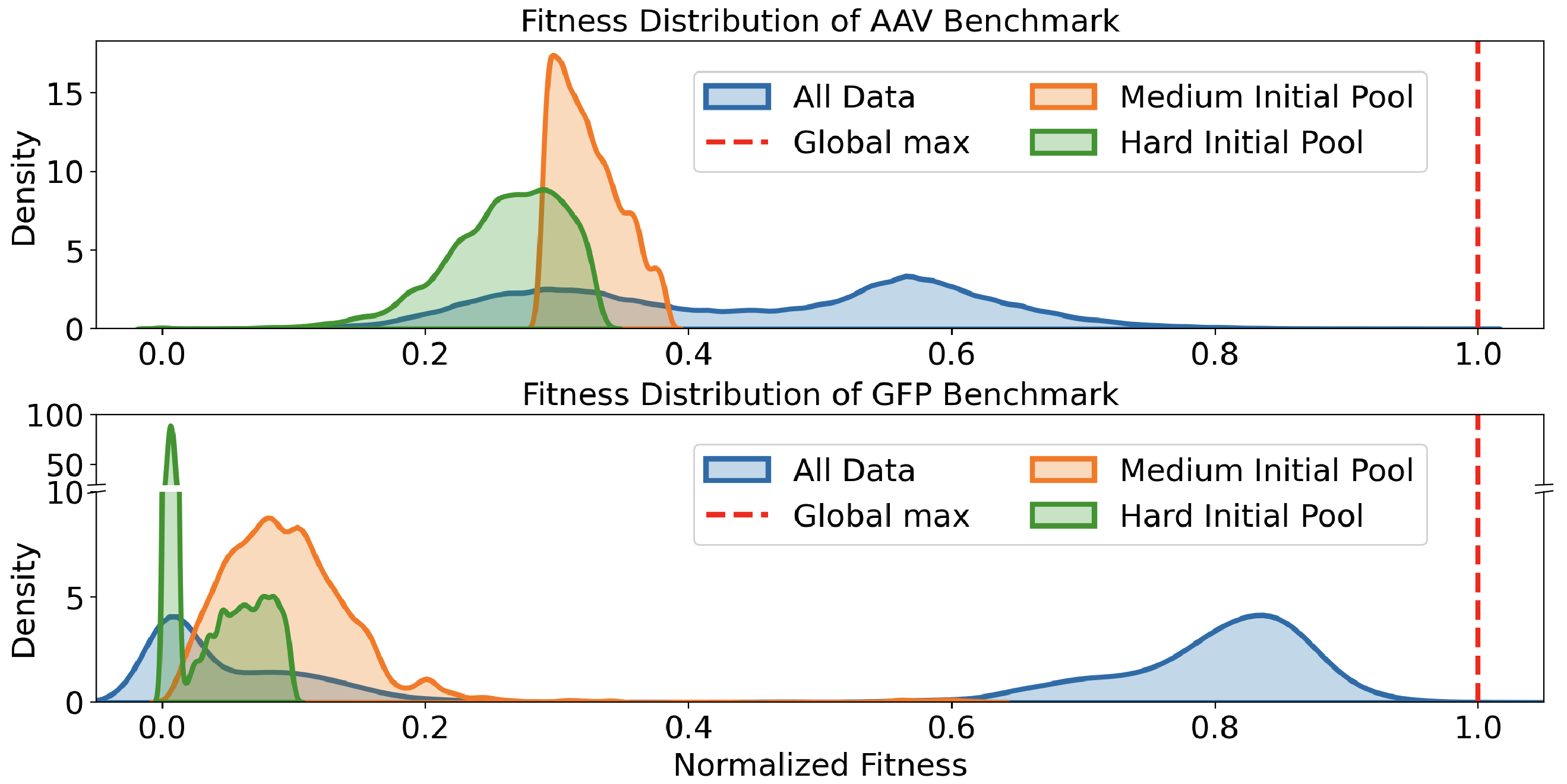}
    \caption{\textbf{Kernel density estimation (KDE) of fitness distributions} for AAV and GFP benchmarks.}
    \label{fig:kde}
\end{figure}

Following the settings of~\citet{kirjner2024ggs}, both GFP and AAV tasks are partitioned into two levels, \textit{medium} and \textit{hard}, by controlling the initial variant pool. For \textit{medium} level, initial mutants are sampled from the \textit{20th--40th} fitness percentiles of the full dataset with a minimum pairwise Hamming distance of 6, resulting in 2,139 and 2,828 initial variants for AAV and GFP, respectively. For \textit{hard} level, initial sequences are drawn from the bottom \textit{30th} percentile with a minimum distance of 7, yielding 3,448 and 2,426 variants for AAV and GFP, respectively. Table~\ref{tab:app-aav-dataset} and~\ref{tab:app-gfp-dataset} describe the statistic of four initial sets. Figure~\ref{fig:kde} provides a visualized comparison on fitness distributions among all data, medium initial sets and hard initial sets.

\begin{table}[h]
    \caption{\textbf{Description of initial sets in the AAV benchmark.} Mean Fitness is the mean normalized fitness of top 128 sequences in subsets.}
    \label{tab:app-aav-dataset}
    \centering
    \begin{tabular}{ccccc}
        \toprule
        Level   & Range(\%) & Gap & Size & Mean Fitness \\
        \midrule
        Medium  & 20-40th         & 6 & 2139 & 0.376 \\
        Hard    & \textless 30th  & 7 & 3448 & 0.326 \\
        \bottomrule
    \end{tabular}
\end{table}

\begin{table}[h]
    \caption{\textbf{Description of initial sets in the GFP benchmark.} Mean Fitness is the mean normalized fitness of top 128 sequences in subsets.}
    \label{tab:app-gfp-dataset}
    \centering
    \begin{tabular}{ccccc}
        \toprule
        Level   & Range(\%) & Gap & Size & Mean Fitness \\
        \midrule
        Medium  & 20-40th         & 6 & 2828 & 0.232 \\
        Hard    & \textless 30th  & 7 & 2426 & 0.092 \\
        \bottomrule
    \end{tabular}
\end{table}

\subsection{Baselines for combinatorial benchmarks}
\label{app:baselines-4site}

We evaluate various baselines on four-site combinatorial benchmarks, GB1 and PhoQ. 

\textbf{CMA-ES}~\citep{hansen2003cmaes}. The Covariance Matrix Adaptation Evolution Strategy optimizes protein sequences through a continuous relaxation of the discrete mutation space, where each sequence is represented by a one-hot encoding and optimized in a continuous domain.

\textbf{BO}~\citep{wilson2017bo}. Bayesian optimization formulates protein directed evolution as a black-box optimization problem by updating an acquisition function after each round. The implementation of BO is derived from FLEXS benchmark~\citep{sinai2020adalead}.

\textbf{AdaLead}~\citep{sinai2020adalead}. This work adopts an adaptive greedy search algorithm for directed evolution by iteratively selecting sequences whose fitness exceeds a dynamically updated threshold based on the previous batch, and generating new candidates via mutation and recombination of these high-performing variants.

\textbf{MLDE}~\citep{wu2019mlde}. This work first incorporates machine learning into the combinatorial directed evolution workflow by training an ensemble model to increase throughput with in-silico modeling.

\textbf{ftMLDE}~\citep{wittmann2021ftmlde}. This work evaluates how different protein encoding methods and training set partitioning strategies affect the performance of protein directed evolution, with EVmutation and MSATransformer adopted as our baselines following~\citet{wang2024knowrlm}.

\textbf{CLADE}~\citep{qiu2021clade}. This work proposes an unsupervised hierarchical clustering strategy to explore the combinatorial mutation space, followed by in silico screening of selected mutants using a supervised proxy model.

\textbf{CLADE 2.0}~\citep{qiu2022clade2}. This work extends CLADE by improving the initial sampling stage through ensembling multiple evolutionary scores, resulting in more robust performance with reduced sensitivity to hyperparameter choices.

\textbf{PEX}~\citep{ren2022pex}. This work introduces proximal exploration, which biases evolutionary search toward high-fitness mutants with low mutation counts by applying a distance-based preference around the wild-type sequence.

\textbf{EvoPlay}~\citep{wang2023evoplay}. This work models protein directed evolution by a neural network-guided Monte Carlo tree search (MCTS) procedure, which is more interpretable than optimization methods that performed in a latent space.

\textbf{KnowRLM}~\citep{wang2024knowrlm}. This work introduces an Amino Acid Knowledge Graph (AAKG) together with a dynamic window mechanism to guide mutation selection using physicochemical properties, optimizing by PPO.

Since MLDE methods~\cite{wu2019mlde, wittmann2021ftmlde, qiu2021clade, qiu2022clade2} need to enumerate all possible mutants to construct a scoring vector, they may not be suitable for full-length benchmarks, where the exponentially large mutation space ($20^L$) makes such enumeration infeasible. We additionally conduct proxy-based search baselines, including random search and brute-force search, to evaluate the contributions of both the proxy model and the search methods; the results are reported in Appendix~\ref{app:isolation}.

\subsection{Baselines for full-length benchmarks}
\label{app:baselines-full-length}

This section introduces additional baselines for full-length benchmarks.

\textbf{GGS}~\citep{kirjner2024ggs}. This work employs a graph-based smoothing approach to train a smoothed proxy model, which defines a discrete energy function for mutant sampling via Gibbs-with-Gradients. Following prior work~\citep{lee2024latprotrl, bogensperger2025vlgpo} on the GFP and AAV benchmarks, we use the provided checkpoints of both the smoothed proxy and the oracle in this work.

\textbf{LatProtRL}~\citep{lee2024latprotrl}. This work encodes protein mutants into latent representations using a pretrained variational encoder–decoder and treats these latent features as states for PPO, with actions defined as small continuous perturbations in the latent space, which improves sample efficiency by reducing the dimension of the mutation space.

\textbf{VLGPO}~\citep{bogensperger2025vlgpo}. This work introduces Variational Latent Generative Protein Optimization, which optimizes the sampling strategy in a continuous latent space using a flow-matching prior. As the original work demonstrates better performance when using an unsmoothed fitness predictor proposed in~\citep{kirjner2024ggs}, we follow this choice and train the model for 15 rounds in an active learning setting.

\textbf{GFN-$\delta$-CS}~\citep{kim2025deltacs}. GFN-AL~\citep{jain2022gflownetsal} leverages Generative Flow Networks (GFlowNets;~\citealp{bengio2021gflownet}) within an active learning loop for protein optimization. This work further improves GFN-AL by introducing a controllable conservation factor $\delta$ to explicitly balance exploration and exploitation. In our experiments, we set $\delta=0.01$ to prioritize fitness optimization.

Since these methods rely on pretraining an encoder–decoder or performing optimization in continuous latent spaces, we carefully considered their compatibility and, for fairness and clarity of comparison, did not include them as baselines for the combinatorial benchmarks.

We additionally evaluate classical proxy-based search methods for full-length benchmarks, including hill climbing, beam search and vanilla Gibbs sampling. The results are reported in Appendix~\ref{app:isolation}. To assess the effectiveness of reference structure input, we also develop simple structure-aware extensions of two strong existing baselines, VLGPO and GFN-$\delta$-CS. Implementation and experiment details are provided in Appendix~\ref{app:structure-baselines}.

\subsection{Metrics for combinatorial benchmarks}
\label{app:metric-4site}

The \textit{mean} and \textit{maximum} fitness are computed over the combined candidate set, which includes 96 mutants from initial set sampled by CLADE~\citep{qiu2021clade} via an unsupervised hierarchical clustering approach, 288 mutants proposed across three rounds ($N=96$ per round), as well as the top 96 mutants selected by the proxy model after the final training round, resulting in a total of 480 candidates.

The normalized discounted cumulative gain (NDCG) is calculated between predictions and the ground truth fitness in the entire mutation space with size of $20^4=160,000$. We compute NDCG using \texttt{metrics.ndcg\_score} implemented in scikit-learn framework~\citep{feurer2015sklearn}.

\subsection{Metrics for full-length benchmarks}
\label{app:metric-full-length}

The \textit{mean} and \textit{maximum} fitness are computed on the top-N of the proposed sequences across 15 rounds, where $N$ is set to 128. The \textit{diversity} of $N$ proposed sequences $\mathcal{D}=\{X_i\}_{i=1}^N$ is defined as mean pairwise Hamming distance~\citep{norouzi2012hamming} among sequences:
\begin{equation}
    Diversity(\mathcal{D})=\frac{1}{N(N-1)}\sum_{i=1}^{N}{\sum_{j=1,j\neq i}^{N}{\texttt{Hamming}(X_i, X_j)}}.
\end{equation}
The \textit{novelty} of $N$ proposed sequences $\mathcal{D}=\{X_i\}_{i=1}^N$ is defined as the mean minimum Hamming distance from each proposed sequence to the initial set $\mathcal{D}_0$:
\begin{equation}
    Novelty(\mathcal{D})=\frac{1}{N}\sum_{i=1}^{N}{\left(\min_{X^*\in \mathcal{D}_0} \left\{\texttt{Hamming}(X_i, X^*)\right\} \right)}.
\end{equation}

\section{Implementation details}
\label{app:implementation}

\subsection{Structural processing}
\label{app:structure-process}

\begin{table}[h]
\centering
\caption{Information of the reference protein structures used in our work.}
\label{tab:app-struc}
\begin{tabular}{cccccc}
    \toprule
    Protein & Source & Structure ID & Chain ID & Position & Length \\
    \midrule
    \textbf{GB1}  & PDB   & \texttt{2GI9}          & A & 1--56    & 56 \\
    \textbf{PhoQ} & AFDB  & \texttt{AF-Q8FIB8-F1-v6}  & A & 188--376 & 189 \\
    \textbf{AAV}  & PDB   & \texttt{3NG9}          & A & 561--588 & 28 \\
    \textbf{GFP}  & PDB   & \texttt{1GFL}          & A & 2--238   & 237 \\
    \bottomrule
\end{tabular}
\end{table}

The sources of protein structures used in this work are listed in Table~\ref{tab:app-struc}. Experimental structures of GB1, AAV and GFP are obtained from the Protein Data Bank (PDB)~\citep{berman2000pdb}, the largest global database of biological macromolecular structures. As the crystal structure of PhoQ is unavailable in the PDB, we select a high-confidence predicted structure with average pLDDT of 82.94 from the AlphaFold Protein Structure Database (AFDB)~\citep{jumper2021afdb} instead.

\subsection{Training details}
\label{app:experimental-settings}

\textbf{Model architecture.} We employed the open-source ESM-2~\citep{lin2022esm2} as PLM for encoding protein sequences, and ESM-IF~\citep{hsu2022esmif} as the inverse folding model for encoding structures. The \textit{ValueNet}, required for computing the value loss in PPO, is implemented as a multi-layer MLP (following KnowRLM~\cite{wang2024knowrlm}). The \textit{ActionNet} (described in Section~\ref{sec:method-hierarchical}) consists of a position action network followed by an amino acid type action network, each is implemented as a two-layer linear network. The \textit{projector}, aiming to transform the sequence-derived delta features into the structural feature space, is implemented as a lightweight linear projection layer. Our code is developed within python 3.11 and PyTorch framework, running on a single NVIDIA A800 80G GPU.

\textbf{Proxy model training details.} For four-site combinatorial benchmarks, we adopt the structure-aware ProSST~\citep{li2024prosst} as our proxy model. ProSST incorporates a structure quantization module and a transformer-based disentangled attention mechanism to better align sequence and structural features. We append a regression head to the backbone, with a hidden size of 128 and a scalar as output, to convert it into a fitness predictor. During all fine-tuning and inference process of ProSST, we utilize the reference structure coordinates as structural inputs. For full-length benchmarks, we use the provided checkpoints of the smoothed proxy in~\citet{kirjner2024ggs} by convention. The proxy is based on a convolutional neural network (CNN) architecture and encodes protein sequences to one-hot embeddings as inputs. For all benchmarks, the proxy model is fine-tuned on the newly proposed mutant candidates for 30 epochs after each round, using a batch size of 96 and a learning rate of $2\times10^{-3}$. The total compute time for an individual experimental run is around 2 hours.

\textbf{RL training details.} The reinforcement learning procedure is optimized using Proximal Policy Optimization (PPO)~\citep{schulman2017ppo}, supplemented by the \texttt{Stable-Baselines3}~\citep{raffin2021sb3} framework. For all benchmarks, we train the PPO using 8 parallel environments, with a clipping range of 0.3, entropy coefficient of 0.0, discount factor of 0.99, learning rate of $3\times10^{-4}$, and batch size of 64. The max episode length $m_{step}$ is set to 3 for each round. The total number of training steps is set to 3,000 for combinatorial benchmarks and 15,000 for full-length benchmarks. For geometric constraint, the weight $\lambda_\text{geo}$ is set to 0.05 and the scaling factor $c$ is set to 1.

\section{Additional experiment results}

\subsection{Full results on 4-site benchmarks}
\label{app:full-results-4site}

We provide results on 3 optimization rounds at Table~\ref{tab:app-full-result-gb1} and Table~\ref{tab:app-full-result-phoq}.

\begin{table}[h]
\caption{\textbf{Results on GB1 benchmarks across 3 rounds}. The best scores \textbf{bolded} and the second best \underline{underlined}. All baseline results are derived from \citet{wang2024knowrlm}.}
\label{tab:app-full-result-gb1}
\centering
\resizebox{1\textwidth}{!}{
\begin{tabular}{lccccccccc}
    \toprule
    & \multicolumn{3}{c}{Round 1} & \multicolumn{3}{c}{Round 2} & \multicolumn{3}{c}{Round 3} \\
    \cmidrule(r){2-4} \cmidrule(r){5-7} \cmidrule(r){8-10}
    Method & 
    Max$\uparrow$ & Mean$\uparrow$ & NDCG$\uparrow$ & 
    Max$\uparrow$ & Mean$\uparrow$ & NDCG$\uparrow$ & 
    Max$\uparrow$ & Mean$\uparrow$ & NDCG$\uparrow$ \\
    \midrule
    MLDE      & 0.650 & 0.183 & 0.767 & 0.680 & 0.217 & 0.794 & 0.684 & 0.203 & 0.789 \\
    ftMLDE (EVmutation)   & 0.725 & 0.233 & 0.791 & 0.770 & 0.280 & 0.814 & 0.935 & 0.414 & 0.833 \\
    ftMLDE (Transformer)  & 0.761 & 0.239 & 0.792 & 0.814 & 0.298 & 0.819 & 0.932 & 0.416 & 0.813  \\
    EvoPlay   & 0.837 & 0.433 & 0.826 & 0.834 & 0.457 & 0.839 & 0.840 & 0.460 & 0.847  \\
    CLADE     & 0.785 & 0.309 & 0.801 & 0.802 & 0.303 & 0.803 & 0.835 & 0.458 & 0.857  \\
    CLADE 2.0  & 0.886 & 0.376 & 0.808 & 0.903 & 0.419 & 0.858 & 0.935 & 0.491 & 0.879 \\
    KnowRLM & \underline{0.931} & \underline{0.494} & \underline{0.851} & \underline{0.972} & \underline{0.534} & \underline{0.862} &  
    \underline{0.972} & \underline{0.562} & \underline{0.884} \\
    \midrule
    \textbf{StructEvo (ours)} & \textbf{0.942} & \textbf{0.514} & \textbf{0.870} & \textbf{0.973} & \textbf{0.540} & \textbf{0.902} &  
    \textbf{0.990} & \textbf{0.587} & \textbf{0.932} \\
    \bottomrule
\end{tabular}}
\end{table}

\begin{table}[h]
\caption{\textbf{Results on PhoQ benchmarks across 3 rounds}. The best scores \textbf{bolded} and the second best \underline{underlined}. All baseline results are derived from \citet{wang2024knowrlm}.}
\label{tab:app-full-result-phoq}
\centering
\resizebox{1\textwidth}{!}{
\begin{tabular}{lccccccccc}
    \toprule
    & \multicolumn{3}{c}{Round 1} & \multicolumn{3}{c}{Round 2} & \multicolumn{3}{c}{Round 3} \\
    \cmidrule(r){2-4} \cmidrule(r){5-7} \cmidrule(r){8-10}
    Method & 
    Max$\uparrow$ & Mean$\uparrow$ & NDCG$\uparrow$ & 
    Max$\uparrow$ & Mean$\uparrow$ & NDCG$\uparrow$ & 
    Max$\uparrow$ & Mean$\uparrow$ & NDCG$\uparrow$ \\
    \midrule
    MLDE      & 0.309 & 0.069 & 0.753 & 0.364 & 0.087 & 0.775 & 0.361 & 0.095 & 0.791 \\
    ftMLDE (EVmutation)   & 0.297 & 0.072 & 0.754 & 0.431 & 0.114 & 0.807 & 0.436 & 0.115 & 0.804 \\
    ftMLDE (Transformer)  & 0.346 & 0.073 & 0.756 & 0.414 & 0.108 & 0.802 & 0.422 & 0.117 & 0.815 \\
    EvoPlay   & 0.443 & 0.119 & 0.782 & 0.444 & 0.135 & 0.786 & 0.474 & 0.143 & 0.804 \\
    CLADE     & 0.319 & 0.070 & 0.759 & 0.441 & 0.089 & 0.762 & 0.467 & 0.089 & 0.777 \\
    CLADE 2.0  & 0.345 & 0.118 & 0.781 & 0.383 & 0.125 & 0.779 & 0.399 & 0.148 & 0.814 \\
    KnowRLM & \underline{0.486} & \underline{0.129} & \underline{0.821} & \underline{0.532} & \underline{0.152} & \underline{0.816} & \textbf{0.658} & \underline{0.157} & \underline{0.819} \\
    \midrule
    \textbf{StructEvo (ours)} & \textbf{0.530} & \textbf{0.158} & \textbf{0.844} & \textbf{0.581} & \textbf{0.169} & \textbf{0.848} & \underline{0.640} & \textbf{0.186} & \textbf{0.856} \\
    \bottomrule
\end{tabular}}
\end{table}

\subsection{Full ablation results on full-length benchmarks}
\label{app:full-ablation}

See Table~\ref{tab:app-full-ablation} for full ablation results. Ablation settings are: (1) removing delta module by directly fusing mutant sequence with the static reference structure via cross-attention, (2) removing structure and fusing two sequence features, (3) relying solely on mutant sequence, and (4) replacing the hierarchical action network with a flat network that jointly samples both actions. The results on the two medium settings show trends consistent with those on the hard setting, as analyzed in Sec.~\ref{sec:exp-ablation}.

\begin{figure}[h]
\centering
\captionof{table}{\textbf{Full results of ablation studies.} w/o: without.}
\label{tab:app-full-ablation}
\begin{tabular}{clcccc}
    \toprule
    &Method    &  Mean$\uparrow$  &   Max$\uparrow$  &  Diversity  &  Novelty \\
    \midrule
    \multirow{5}{1cm}{\centering \emph{AAV}\\ \emph{medium}}
    &\textbf{StructEvo} & \textbf{0.82}$\pm$0.01 & \textbf{0.86}$\pm$0.01 
                        & 4.5$\pm$0.7 & 8.6$\pm$0.2 \\
    &(1) w/o delta      & 0.79$\pm$0.00 & 0.83$\pm$0.01 & 5.5$\pm$0.8 & 7.7$\pm$0.0 \\
    &(2) w/o structure  & 0.77$\pm$0.01 & 0.81$\pm$0.02 & 4.7$\pm$1.3 & 8.1$\pm$0.7 \\
    &(3) w/o reference  & 0.77$\pm$0.01 & 0.81$\pm$0.02 & 4.8$\pm$2.0 & 7.8$\pm$0.0 \\
    &(4) w/o hierarchy  & 0.78$\pm$0.01 & 0.80$\pm$0.01 & 5.3$\pm$1.3 & 7.9$\pm$0.6 \\
    \midrule
    \multirow{5}{1cm}{\centering \emph{AAV}\\ \emph{hard}} 
    &\textbf{StructEvo} & \textbf{0.83}$\pm$0.03 & \textbf{0.87}$\pm$0.04  
                        &  4.5$\pm$0.3 & 8.4$\pm$0.5 \\
    &(1) w/o delta      &  0.80$\pm$0.03 &  0.84$\pm$0.05 & 4.9$\pm$0.6 & 8.3$\pm$0.7 \\
    &(2) w/o structure  &  0.75$\pm$0.03 &  0.79$\pm$0.04 & 4.9$\pm$1.8 & 8.6$\pm$0.7 \\
    &(3) w/o reference  &  0.75$\pm$0.01 &  0.81$\pm$0.01 & 7.5$\pm$1.4 & 8.7$\pm$0.4 \\
    &(4) w/o hierarchy  &  0.77$\pm$0.02 &  0.81$\pm$0.02 & 5.2$\pm$1.4 & 7.9$\pm$0.3 \\
    \midrule
    \multirow{5}{1cm}{\centering \emph{GFP}\\ \emph{medium}}
    &\textbf{StructEvo} & \textbf{1.11}$\pm$0.02 & \textbf{1.17}$\pm$0.02 
                        & 6.3$\pm$0.8 & 11.9$\pm$0.8 \\
    &(1) w/o delta      & 1.09$\pm$0.02 & 1.13$\pm$0.02 & 6.2$\pm$1.3 & 10.9$\pm$1.1 \\
    &(2) w/o structure  & 1.05$\pm$0.02 & 1.09$\pm$0.02 & 7.4$\pm$1.7 & 10.7$\pm$0.5 \\
    &(3) w/o reference  & 1.06$\pm$0.05 & 1.10$\pm$0.05 & 6.2$\pm$1.6 & 11.5$\pm$1.6 \\
    &(4) w/o hierarchy  & 0.91$\pm$0.01 & 0.98$\pm$0.01 & 10.5$\pm$2.5 & 7.4$\pm$1.4 \\
    \midrule
    \multirow{5}{1cm}{\centering \emph{GFP}\\ \emph{hard}}
    &\textbf{StructEvo} & \textbf{1.00}$\pm$0.04 & \textbf{1.05}$\pm$0.05  
                        & 4.9$\pm$0.3 & 11.4$\pm$0.6 \\
    &(1) w/o delta      & 0.97$\pm$0.05 & 1.01$\pm$0.05 & 4.7$\pm$0.3 & 11.0$\pm$1.1 \\
    &(2) w/o structure  & 0.83$\pm$0.04 & 0.89$\pm$0.02 & 5.7$\pm$1.2 & 11.7$\pm$0.8 \\
    &(3) w/o reference  & 0.84$\pm$0.06 & 0.93$\pm$0.05 & 5.9$\pm$0.9 & 11.3$\pm$1.2 \\
    &(4) w/o hierarchy  & 0.75$\pm$0.02 & 0.82$\pm$0.01 & 13.8$\pm$2.2 & 13.0$\pm$1.8 \\
    \bottomrule
\end{tabular}
\end{figure}

\section{Additional in-depth analysis}
\label{app:additional-indepth}

We discuss additional ablation effects in this section, including analysis on structure prediction models (Sec.~\ref{app:protenix}), proxy model robustness (Sec.~\ref{app:proxy-robust}), and weights for geometric constraint (Sec.~\ref{app:geo_coef}).

\subsection{Analysis on structure prediction models}
\label{app:protenix}

In this section, we assess the sensitivity of structure prediction models to point mutations on GFP mutants. We choose Protenix~\cite{bytedance2025protenix} as the representative structure prediction model for assessment. Compared to recent well-known models such as AlphaFold3, Protenix is completely open-source while achieving comparable performance across multiple structure prediction benchmarks.

Specifically, we perform stratified sampling on the GFP dataset by partitioning mutant sequences into 10 bins according to their fitness values, and randomly selecting 10 mutant sequences from each bin, resulting in a total of 100 sequences. The average Hamming distance of these 100 sequences is 7.5, indicating a moderate level of mutational diversity for evaluating model sensitivity.

For each sequence, we predict its structure using the default Protenix inference configuration, including MSA construction via UniRef100, 5 Pairformer cycles, and 200 diffusion steps. We adopt the \texttt{protenix\_base\_default\_v1.0.0} version and run all experiments on a single NVIDIA A800 80GB GPU. Due to the expensive computational costs, we only generate one structure sample per sequence. The entire MSA search and structure prediction process for 100 sequences takes approximately 13 hours, making full-scale structure prediction prohibitive within the RL training loop, which requires evaluating substantially more candidates.

Under this setting, we evaluate two aspects. First, we examine the Pearson and Spearman correlations (implemented by \texttt{scipy.stats}) between three commonly used confidence metrics and fitness values:
\begin{itemize}
    \item \textit{pLDDT (predicted Local Distance Difference Test)}: Higher pLDDT values indicate greater confidence in local structural predictions at the residue level.
    \item \textit{GpDE (Global predicted Distance Error)}: Lower distance errors indicate more reliable predictions. Therefore, we use the negative values of GpDE scores to compute correlations.
    \item \textit{pTM (predicted TM-score)}: Values closer to 1 indicate higher confidence in the overall structural prediction, capturing the reliability of global structure.
\end{itemize}

\begin{figure}[h]
    \centering
    \includegraphics[width=0.8\linewidth]{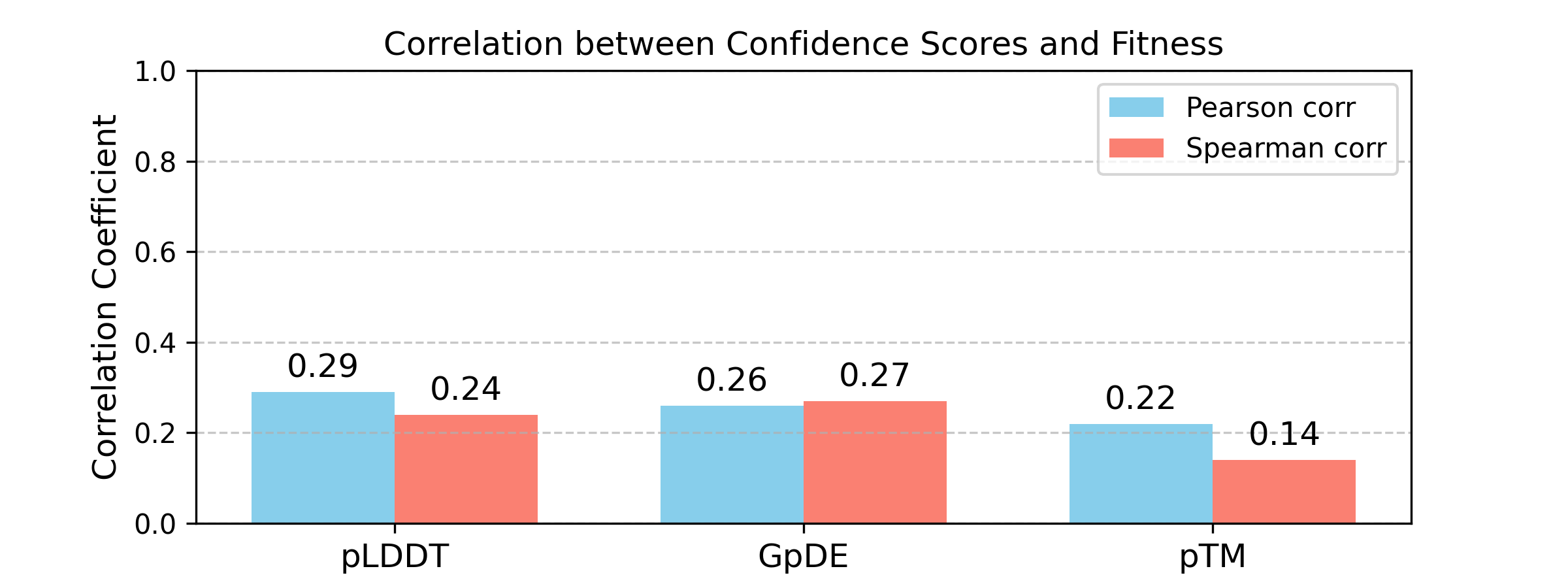}
    \caption{Correlation between Confidence Scores and Fitness.}
    \label{fig:conf-corr}
\end{figure}

The results are shown in Figure~\ref{fig:conf-corr}. All three confidence metrics exhibit consistently weak correlations with fitness, indicating a poor alignment between structure prediction confidence and protein functionality. Notably, the highest observed correlation (0.29 for pLDDT-Pearson) remains below 0.3, suggesting that confidence scores are not sufficiently sensitive to the subtle structural variations induced by point mutations.

To further investigate the sensitivity of predicted structures, we conduct two analyses:
\begin{itemize}
    \item \textit{Reference-based difference}: We calculate the structure deviation between the 100 predicted mutant structures with the reference structure, and
    examine the correlation between the structural deviation and fitness value.
    \item \textit{Pairwise difference}: For each pair in the 100 predicted mutant structures (a total of 4950 pairs), we examine the correlation between structural difference and the corresponding fitness difference. The fitness difference is defined as the absolute fitness gap.
\end{itemize}
In both analyses, the structural difference is measured by $C\alpha$-RMSD (root mean square deviation) with rigid-body alignment implemented by the \texttt{BioPython.Superimposer} module. 

\begin{figure}[h]
    \centering
    \includegraphics[width=0.9\linewidth]{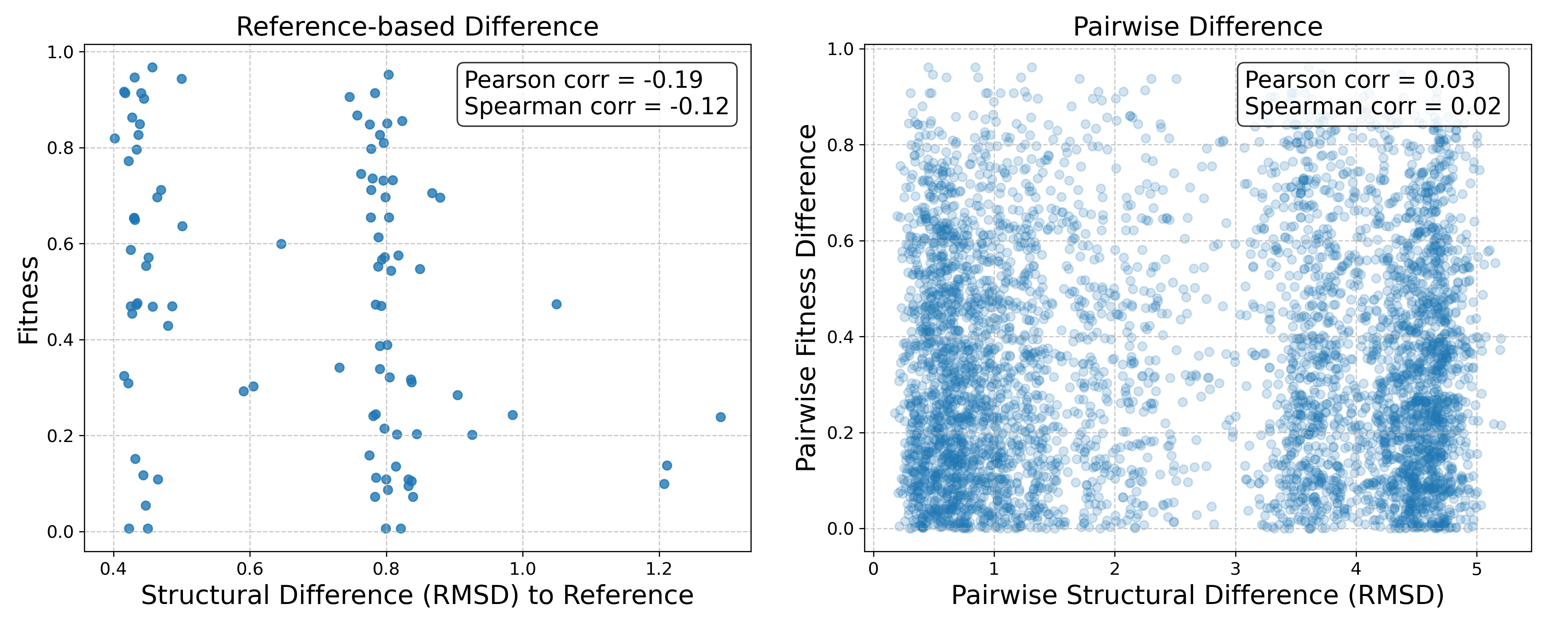}
    \caption{Samples of reference-based difference (left) and pairwise difference (right).}
    \label{fig:diff}
\end{figure}

The results are shown in Figure~\ref{fig:diff}. We find that both structural differences reveal negligible relationships with fitness values. In the reference-based setting, structural deviation from the reference shows weak negative correlations with fitness (Pearson -0.19, Spearman -0.12). In the pairwise setting, the correlation is much lower (Pearson 0.03, Spearman 0.02). The diffuse scatter distributions further indicate weak relationship between structural variations and functional changes. Notably, these correlations are even weaker than those of confidence scores. This suggests that current structure prediction outputs yield weak relationships to point mutations, limiting their utility for modeling fine-grained fitness landscape and guiding protein optimization. 

Given the high computational costs of structure prediction, a potential alternative strategy is to periodically update the reference structure with newly predicted mutant structure, for example every $T$ steps or when the accumulated mutation distance exceeds a threshold. However, this approach requires careful evaluation of update schedules. We leave this exploration to future work.

\subsection{Robustness analysis of the proxy model}
\label{app:proxy-robust}

To examine the robustness of proxy model, we conduct experiments on the GFP-hard task under several settings:
\begin{itemize}
    \item \textit{Default}: The proxy model provides rewards during RL. At each round, the proxy selects top-N candidates, which are then evaluated by the oracle and used to iteratively refine the proxy.
    \item \textit{Oracle-full}: The proxy is replaced by the oracle to provide rewards and can be queried without constraint. This represents an ideal upper bound of the proxy.
    \item \textit{Oracle-limited}: The proxy is replaced by the oracle, but the number of queries is strictly limited as default setting (i.e., N per round).
\end{itemize}

The results are listed below:

\begin{table}[h]
    \centering
    \caption{Proxy robustness analysis on GFP-hard task.}
    \label{tab:proxy-robust}
    \begin{tabular}{lcccc}
    \toprule
                            & Mean$\uparrow$ & Max$\uparrow$  &  Diversity  &  Novelty \\
    \midrule
    \textit{Default}        & 1.00$\pm$0.04 & 1.05$\pm$0.05 &  4.9$\pm$0.3 & 11.4$\pm$0.6\\
    \textit{Oracle-full}    & 1.07$\pm$0.01 & 1.12$\pm$0.02 & 12.9$\pm$2.9 & 13.1$\pm$1.2 \\
    \textit{Oracle-limited} & 0.41$\pm$0.03 & 0.49$\pm$0.04 & 16.7$\pm$4.2 &  5.8$\pm$0.9\\
    \bottomrule
    \end{tabular}
\end{table}

We observe that \textit{oracle-full} achieves the best performance as an idealized upper bound setting. However, \textit{this setting is impractical in MLDE, as oracle queries correspond to costly wet-lab experiment budgets}. In contrast, \textit{oracle-limited} shows a significant performance drop, indicating that the RL agent collects insufficient trajectories under limited signals. This further demonstrates that the frequent rewards from the proxy model are crucial for effective evolution. We will include a more comprehensive analysis of proxy effects on different rounds and tasks in future work.

\subsection{Isolation of proxy and RL policy}
\label{app:isolation}

To isolate the contribution of RL policy and proxy model, we conduct the following proxy-based searching methods as additional baselines:

\begin{itemize}
    \item \textbf{Random}: randomly select single mutations each round without using the proxy.
    \item \textbf{Brute-force} (4-site benchmarks): rank all 160,000 candidates by proxy and select top N=96 for each round.
    \item \textbf{Hill climbing} (full-length benchmarks): use proxy to rank all single-mutation neighbors of each current sequence, and greedily select the best N candidates for the next round.
    \item \textbf{Beam search} (full-length benchmarks): use proxy to rank all single-mutation neighbors, while keeping a beam width of M=5 paths for each sequence.
    \item \textbf{Gibbs sampling} (full-length benchmarks): we conduct a \textit{vanilla} Gibbs sampling that uses proxy to rank all single-mutation neighbors and sample new candidates from a Boltzmann distribution. One of our original baselines, GGS, is also an expansion of Gibbs sampling that adds graph-based smoothing.
\end{itemize}

For each method, we conduct 2 experimental settings:

\begin{itemize}
    \item \textbf{Full-round}: Starting from the initial pool, we update the proxy using candidates selected by each method. We report performance at R rounds (R=3 for 4-site and R=15 for full-length).
    \item \textbf{Fixed-proxy}: We fix the fine-tuned proxy model at the last round in our main experiment and only perform the last round. This comparison can directly compare the effectiveness of each search strategy based on the same proxy model.
\end{itemize}

Each baseline is evaluated over three independent runs. Both average results and standard deviations are reported in Table~\ref{tab:app-iso-4site} and ~\ref{tab:app-iso-full}.

\begin{table}[h]
\centering
\caption{Comparison with proxy-based search baselines on 4-site benchmarks.}
\label{tab:app-iso-4site}
\resizebox{\linewidth}{!}{
\begin{tabular}{lcccccc}
\toprule
\multirow{2}{*}{Method} & \multicolumn{3}{c}{GB1} & \multicolumn{3}{c}{PhoQ} \\
\cmidrule(lr){2-4} \cmidrule(lr){5-7}
& Mean & Max & NDCG & Mean & Max & NDCG \\
\midrule
Random
& 0.08$\pm$0.00 & 0.22$\pm$0.12 & -
& 0.00$\pm$0.00 & 0.14$\pm$0.02 & - \\
Brute-force (full-round)
& 0.37$\pm$0.01 & 0.60$\pm$0.01 & 0.78$\pm$0.01
& 0.10$\pm$0.01 & 0.23$\pm$0.02 & 0.71$\pm$0.01 \\
Brute-force (fixed-proxy)
& 0.45$\pm$0.01 & 0.97$\pm$0.02 & -
& 0.09$\pm$0.01 & 0.55$\pm$0.12 & - \\
\midrule
\textbf{StructEvo (ours)}
& \textbf{0.59$\pm$0.01} & \textbf{0.99$\pm$0.01} & \textbf{0.93$\pm$0.01}
& \textbf{0.19$\pm$0.02} & \textbf{0.64$\pm$0.07} & \textbf{0.86$\pm$0.01} \\
\bottomrule
\end{tabular}
}
\end{table}

\begin{table*}[h]
\centering
\caption{Comparison with proxy-based search baselines on full-length benchmarks.}
\label{tab:app-iso-full}
\resizebox{\textwidth}{!}{
\begin{tabular}{lcccccccc}
\toprule
\multirow{2}{*}{Method}
& \multicolumn{4}{c}{AAV-hard}
& \multicolumn{4}{c}{GFP-hard} \\
\cmidrule(lr){2-5} \cmidrule(lr){6-9}
& Mean & Max & Diversity & Novelty
& Mean & Max & Diversity & Novelty \\
\midrule
Random
& 0.08$\pm$0.01 & 0.22$\pm$0.02 & 22.7$\pm$0.3 & 11.3$\pm$0.3
& 0.04$\pm$0.05 & 0.24$\pm$0.07 & 39.6$\pm$1.2 & 14.5$\pm$0.1 \\
Hill climbing (full-round)
& 0.56$\pm$0.03 & 0.68$\pm$0.03 & 4.9$\pm$0.2 & 7.7$\pm$0.2
& 0.68$\pm$0.03 & 0.77$\pm$0.03 & 4.8$\pm$0.3 & 7.5$\pm$0.4 \\
Beam search (full-round)
& 0.62$\pm$0.03 & 0.68$\pm$0.03 & 4.2$\pm$0.2 & 7.9$\pm$0.3
& 0.76$\pm$0.03 & 0.78$\pm$0.03 & 4.4$\pm$0.2 & 9.3$\pm$0.4 \\
Gibbs sampling (full-round)
& 0.51$\pm$0.03 & 0.60$\pm$0.03 & 12.9$\pm$0.3 & 8.0$\pm$0.4
& 0.43$\pm$0.03 & 0.47$\pm$0.03 & 15.8$\pm$0.3 & 13.1$\pm$0.6 \\
Hill climbing (fixed-proxy)
& 0.72$\pm$0.00 & 0.77$\pm$0.01 & 5.8$\pm$0.1 & 7.9$\pm$0.1
& 0.91$\pm$0.00 & 0.95$\pm$0.01 & 4.8$\pm$0.2 & 11.7$\pm$0.1 \\
Beam search (fixed-proxy)
& 0.72$\pm$0.01 & 0.79$\pm$0.01 & 5.7$\pm$0.1 & 7.8$\pm$0.1
& 0.92$\pm$0.00 & 0.95$\pm$0.01 & 4.6$\pm$0.2 & 11.0$\pm$0.1 \\
Gibbs sampling (fixed-proxy)
& 0.69$\pm$0.03 & 0.72$\pm$0.03 & 6.6$\pm$0.2 & 8.1$\pm$0.4
& 0.87$\pm$0.02 & 0.97$\pm$0.04 & 6.1$\pm$0.2 & 12.4$\pm$0.4 \\
\midrule
\textbf{StructEvo (ours)}
& \textbf{0.83$\pm$0.03} & \textbf{0.87$\pm$0.04} & 4.5$\pm$0.3 & 8.4$\pm$0.5
& \textbf{1.00$\pm$0.04} & \textbf{1.05$\pm$0.05} & 4.9$\pm$0.3 & 11.4$\pm$0.6 \\
\bottomrule
\end{tabular}
}
\end{table*}

The results show that:
(1) On the 4-site benchmarks, exhaustive brute-force method selects the global optima under the proxy model, yet still yields lower oracle fitness than StructEvo. This provides direct evidence of prediction errors, where simply maximizing the proxy predicted fitness does not necessarily lead to better navigation on the true landscape. In contrast, \textit{the RL search strategy induces a more constrained search distribution that partly mitigates proxy overestimation}.

(2) We observe the same trend on the full-length tasks. StructEvo consistently outperforms all three simple search baselines, further indicating that the RL policy makes an independent contribution beyond proxy quality. Notably, Gibbs sampling achieves the highest diversity but lower fitness, suggesting that much of this exploration is of low value although it explores a broader region of sequence space. In contrast, \textit{StructEvo achieves a more effective balance between exploration and exploitation}.

(3) Performance drops substantially in the full-round setting compared with the fixed-proxy setting. This is further consistent with our observation, that \textit{policy improvement and proxy refinement are mutually complementary in the active-learning loop}. The RL policy collects more informative candidates to fine-tune a better proxy, while the refined proxy provides more reliable guidance for RL search.

\subsection{Comparison of structure-aware baselines}
\label{app:structure-baselines}

To evaluate more directly the advantage of the reference structure inputs, we implemented simple structure-aware extensions of two strong existing baselines, VLGPO and GFN-$\delta$-CS, providing them with the same per-residue reference structure feature encoding by ESM-IF as used in StructEvo:

\begin{itemize}
    \item \textbf{VLGPO + Struct}: the structural features are normalized and used as weight for the original fitness guidance loss during sampling.
    \item \textbf{GFN-$\delta$-CS + Struct}: the structural features are normalized and used as weight for the mutation action logits produced by the original LSTM network.
\end{itemize}

We note that there does not exist mature peer-reviewed structure-aware MLDE baselines for fair comparison. A few of MLDE studies involve protein structural information, but they are neither evaluated on standard benchmarks nor provide reproducible code. To the best of our knowledge, StructEvo is the first structure-aware RL-based method on MLDE.

\begin{table*}[h]
\centering
\caption{Comparison with structure-aware baselines on full-length benchmarks.}
\label{tab:structure_baselines}
\resizebox{\textwidth}{!}{
\begin{tabular}{lcccccccc}
\toprule
\multirow{2}{*}{Method}
& \multicolumn{4}{c}{AAV-hard}
& \multicolumn{4}{c}{GFP-hard} \\
\cmidrule(lr){2-5} \cmidrule(lr){6-9}
& Mean & Max & Diversity & Novelty
& Mean & Max & Diversity & Novelty \\
\midrule
VLGPO
& 0.71$\pm$0.01 & 0.78$\pm$0.01 & 3.3$\pm$0.2 & 6.9$\pm$0.1
& 0.79$\pm$0.02 & 0.95$\pm$0.00 & 1.9$\pm$0.0 & 6.2$\pm$0.0 \\
VLGPO + Struc
& 0.72$\pm$0.01 & 0.79$\pm$0.00 & 3.3$\pm$0.1 & 7.0$\pm$0.1
& 0.81$\pm$0.02 & 0.95$\pm$0.03 & 1.8$\pm$0.0 & 6.2$\pm$0.1 \\
GFN-$\delta$-CS
& 0.76$\pm$0.03 & 0.80$\pm$0.03 & 4.0$\pm$0.7 & 6.7$\pm$1.1
& 0.86$\pm$0.06 & 0.90$\pm$0.05 & 4.3$\pm$1.6 & 8.2$\pm$1.5 \\
GFN-$\delta$-CS + Struc
& 0.77$\pm$0.06 & 0.84$\pm$0.08 & 5.8$\pm$0.2 & 7.1$\pm$2.5
& 0.84$\pm$0.04 & 0.87$\pm$0.05 & 6.4$\pm$1.9 & 13.7$\pm$3.8 \\
\midrule
\textbf{StructEvo (ours)}
& $\mathbf{0.83\pm0.03}$ & $\mathbf{0.87\pm0.04}$ & 4.5$\pm$0.3 & 8.4$\pm$0.5
& $\mathbf{1.00\pm0.04}$ & $\mathbf{1.05\pm0.05}$ & 4.9$\pm$0.3 & 11.4$\pm$0.6 \\
\bottomrule
\end{tabular}
}
\end{table*}

The results are listed in Table~\ref{tab:structure_baselines}. We observe that StructEvo still consistently achieves the highest mean and maximum fitness on both benchmarks, indicating that \textit{the improvement cannot be attributed merely to the additional input structure, but also arise from how structural guidance is integrated into the RL search policy}. Besides, incorporating structure information into two baselines yields modest performance gains in most cases, while the slight drop of GFN-$\delta$-CS on GFP-hard benchmark may be associated with broader exploration of simple position-wise logits reweighting. More robust integration of structural knowledge could be explored in future work.

\subsection{Single-mutation settings on 4-site benchmark}
\label{app:single-setting}

To explore the purely additive setting, which is more common in real-world protein engineering, we conducted the \textit{single-mutation initial library experiments} on the 4-site GB1 benchmark. The initial library is reconstructed to all 19×4=76 single mutants, with mean and maximum fitness of 0.07 and 0.44, respectively. Accordingly, we set the oracle budget (i.e., the number of proposed candidates per round) to N=76 instead of 96. We evaluated two strong baselines, KnowRLM and CLADE 2.0, as well as the following additive baselines, which sample new candidates from the single-mutation neighbors of current sequences at each round:

\begin{itemize}
    \item \textbf{Random}: randomly selects N=76 single-mutation neighbors per round without proxy scoring.
    \item \textbf{Hill climbing}: greedily chooses mutants with the highest proxy scores.
    \item \textbf{Beam search}: maintains K=5 mutants with top proxy scores and randomly selects one for the next round.
    \item \textbf{Gibbs sampling}: samples new candidates from a Boltzmann distribution of proxy scores.
\end{itemize}

The results are summarized in Table~\ref{tab:gb1_single}:

\begin{table*}[h]
\centering
\caption{Results of single-mutation settings on GB1 benchmark.}
\label{tab:gb1_single}
\resizebox{\textwidth}{!}{
\begin{tabular}{lccccccccc}
\toprule
\multirow{3}{*}{Method} & \multicolumn{3}{c}{Round 1} & \multicolumn{3}{c}{Round 2} & \multicolumn{3}{c}{Round 3} \\
\cmidrule(lr){2-4}\cmidrule(lr){5-7}\cmidrule(lr){8-10}
 & Max & Mean & NDCG & Max & Mean & NDCG & Max & Mean & NDCG \\
\midrule
Random 
 & 0.44$\pm$0.02 & 0.07$\pm$0.00 & - 
 & 0.44$\pm$0.02 & 0.07$\pm$0.01 & - 
 & 0.44$\pm$0.02 & 0.07$\pm$0.01 & - \\
Hill 
 & 0.49$\pm$0.02 & 0.18$\pm$0.00 & 0.73$\pm$0.00 
 & 0.52$\pm$0.04 & 0.19$\pm$0.01 & 0.75$\pm$0.00 
 & 0.58$\pm$0.01 & 0.21$\pm$0.01 & 0.76$\pm$0.00 \\
Beam
 & 0.47$\pm$0.00 & 0.18$\pm$0.00 & 0.76$\pm$0.00 
 & 0.48$\pm$0.03 & 0.23$\pm$0.01 & 0.76$\pm$0.01 
 & 0.49$\pm$0.03 & 0.24$\pm$0.01 & 0.76$\pm$0.01 \\
Gibbs
 & 0.49$\pm$0.03 & 0.21$\pm$0.02 & 0.76$\pm$0.00 
 & 0.54$\pm$0.06 & 0.26$\pm$0.05 & 0.77$\pm$0.01 
 & 0.58$\pm$0.00 & 0.29$\pm$0.04 & 0.78$\pm$0.01 \\
CLADE 2.0 
 & \textbf{0.55$\pm$0.07} & 0.18$\pm$0.01 & \textbf{0.77$\pm$0.01} 
 & \textbf{0.60$\pm$0.11} & 0.23$\pm$0.03 & 0.78$\pm$0.01 
 & \textbf{0.70$\pm$0.09} & 0.27$\pm$0.01 & 0.80$\pm$0.01 \\
KnowRLM 
 & \textbf{0.51$\pm$0.03} & 0.24$\pm$0.04 & \textbf{0.78$\pm$0.01} 
 & \textbf{0.62$\pm$0.02} & 0.33$\pm$0.03 & 0.78$\pm$0.00 
 & 0.66$\pm$0.04 & 0.36$\pm$0.01 & 0.80$\pm$0.00 \\
 \midrule
\textbf{StructEvo}
 & \textbf{0.55$\pm$0.02} & \textbf{0.29$\pm$0.00} & \textbf{0.77$\pm$0.00} 
 & \textbf{0.66$\pm$0.02} & \textbf{0.39$\pm$0.01} & \textbf{0.80$\pm$0.00} 
 & \textbf{0.72$\pm$0.00} & \textbf{0.42$\pm$0.01} & \textbf{0.85$\pm$0.01} \\
\bottomrule
\end{tabular}
}
\end{table*}

We observe that, under the single-mutation initialization setting, StructEvo still outperforms or achieves comparable performance to all baselines across all rounds, improving mean fitness from 0.29 to 0.42. It also achieves the best proxy model ranking quality with an NDCG of 0.85, surpassing KnowRLM and CLADE 2.0 (both 0.80). 

The additive baselines reach limited improvements, as their heavy reliance on proxy scores leads to proxy bias and inaccurate local ranking. In contrast, StructEvo learns an adaptive mutation policy from historical evolutionary trajectories, which enables a better balance between exploration of promising regions and exploitation of high-fitness mutants. Note that the identical results of Random baseline across rounds are due to the cumulative evaluation, where the randomly sampled triple and quadruple mutants are worse than single or double mutants. This further demonstrates the scarcity of valuable mutants in the high dimensional mutation space.

\subsection{Additional case studies of delta features}
\label{app:more-case}

Our ablations and case studies have shown that the delta module improves MLDE performance. To further validate this approximation, we additionally collected several GFP mutants with experimentally resolved or high-confidence predicted structures, as listed in Table~\ref{tab:more-case}. For each mutant, we leveraged a structure encoder (ESM-IF in our work) to compute $h_\text{true}$ (the feature of the true mutant structure from PDB or AFDB), $h_\text{ref}$ (the feature of the reference structure), and $h_\text{approx}$(the approximated structural feature encoded by the delta-structure fusion encoder, i.e., $\hat{h}_\text{str}$ in Section~\ref{sec:method-delta}), and then report:

\begin{itemize}
    \item $Dist.$: Levenshtein distance of amino acid mutations from mutant to reference sequence.
    \item $Sim_\text{r}$: cosine similarity of $h_\text{true}$ and $h_\text{ref}$.
    \item $Sim_\text{a}$: cosine similarity of $h_\text{true}$ and $h_\text{approx}$.
    \item $Sim_\text{d}$: cosine similarity of ($h_\text{true}-h_\text{ref}$) and ($h_\text{approx}-h_\text{ref}$).
\end{itemize}

\begin{table}[h]
    \centering
    \caption{More case study of delta features.}
    \label{tab:more-case}
    \begin{tabular}{lp{5.5cm}cccc}
    \toprule
    Mutant ID & Source & $Dist.$ & $Sim_\text{r}$ & $Sim_\text{a}$ & $Sim_\text{d}$ \\
    \midrule
    \texttt{5N9O} & PDB (experimentally resolved) & 5 & 0.69 & 0.95 & 0.97 \\
    \texttt{8DPD} & PDB (experimentally resolved) & 17 & 0.73 & 0.85 & 0.96 \\
    \texttt{AF-Q8WTD0-F1} & AFDB (AF2 predicted, pLDDT=96.88) & 43 & 0.26 & 0.77 & 0.84 \\
    \bottomrule
    \end{tabular}
\end{table}

Across all three cases, the true mutant structural representations are substantially closer to the approximated mutant representations than the reference. The high Delta Similarity $Sim_\text{d}$ (0.84-0.97) further indicates that the projected delta structure yields meaningful structural changes.

Instead of reconstructing coordinate-level mutant structures, the goal of delta feature is to incorporate mutation-associated structural signals inferred from sequence differences \textit{into the end-to-end optimization process}, especially under MLDE settings where mutant structures are unavailable. In this way, the model directly learns how these structural signals provide guidance for navigating the fitness landscape and therefore selecting high-fitness variants.

\subsection{Sensitivity analysis of the geometric constraint weight}
\label{app:geo_coef}

As introduced in Sec.~\ref{sec:method-geoloss}, the overall RL loss consists of PPO loss and geometric constraint loss:

\begin{equation}
    \mathcal{L}_\text{total} = \underbrace{
        \mathcal{L}_\text{policy}
        + \lambda_\text{entropy} \cdot \mathcal{L}_\text{entropy}
        + \lambda_\text{value} \cdot \mathcal{L}_\text{value}
    }_\text{standard PPO loss}
    + \underbrace{\lambda_\text{geo} \cdot \mathcal{L}_\text{geo}}_\text{geometric constraint}.
\end{equation}

To investigate the effect of the geometric constraint, we vary its weight $\lambda_\text{geo}$ and conduct experiments on GFP-hard task. Each setting is evaluated over 5 independent runs. The results are shown in Table~\ref{tab:geo_coef}.

\begin{table}[h]
    \centering
    \caption{Hyperparameter analysis of geometric constraint weight on GFP-hard task.}
    \label{tab:geo_coef}
    \begin{tabular}{lcccc}
    \toprule
                & Mean$\uparrow$ & Max$\uparrow$ &  Diversity  &  Novelty \\
    \midrule
    $\lambda_\text{geo}=0$ & 
            0.984$\pm$0.081 & 1.017$\pm$0.072 & 5.101$\pm$1.087 & 9.648$\pm$1.795  \\
    $\lambda_\text{geo}=0.01$ & 
            0.987$\pm$0.052 & 1.031$\pm$0.036 & 4.878$\pm$0.312 & \textbf{11.427}$\pm$1.134  \\
    $\lambda_\text{geo}=0.05$ (default) & 
            \textbf{0.998}$\pm$0.041 & \textbf{1.047}$\pm$0.046 & 4.868$\pm$0.336 & 11.425$\pm$0.602  \\
    $\lambda_\text{geo}=0.1$ & 
            0.970$\pm$0.041 & 1.010$\pm$0.043 & 5.289$\pm$0.358 & 10.596$\pm$0.846  \\
    $\lambda_\text{geo}=0.5$ & 
            0.925$\pm$0.043 & 0.991$\pm$0.031 & \textbf{6.452}$\pm$1.534 & 10.677$\pm$0.785  \\
    $\lambda_\text{geo}=1$ & 
            0.931$\pm$0.034 & 0.974$\pm$0.027 & 5.115$\pm$0.970 & 9.706$\pm$0.115  \\
    \bottomrule
    \end{tabular}
\end{table}

We observe that: (1) increasing $\lambda_\text{geo}$ generally improves training stability, as reflected by reduced standard deviations across mean and max fitness metrics, indicating more stable optimization. (2) $\lambda_\text{geo}=0.05$ achieves the best performance on fitness metrics, while both larger and smaller values degrade performance. This is likely because larger weights overly restrict policy updates, whereas smaller weights provide insufficient regularization, leading to noisier gradients and less stable optimization. (3) We found that novelty decreases with larger $\lambda_\text{geo}$, likely because stronger constraint limits exploration and favors more proximal candidates. Overall, we select $\lambda_\text{geo}=0.05$ as the default setting reported in the main experiments.

\newpage
\section*{NeurIPS Paper Checklist}

\begin{enumerate}

\item {\bf Claims}
    \item[] Question: Do the main claims made in the abstract and introduction accurately reflect the paper's contributions and scope?
    \item[] Answer: \answerYes{}
    \item[] Justification: We clearly state the scope and contributions of both communities of protein optimization and computer science in the abstract and introduction.

\item {\bf Limitations}
    \item[] Question: Does the paper discuss the limitations of the work performed by the authors?
    \item[] Answer: \answerYes{}
    \item[] Justification: We discuss limitations at Sec.~\ref{sec:conclusion} .

\item {\bf Theory assumptions and proofs}
    \item[] Question: For each theoretical result, does the paper provide the full set of assumptions and a complete (and correct) proof?
    \item[] Answer: \answerNA{}
    \item[] Justification: The paper does not include theoretical results.

    \item {\bf Experimental result reproducibility}
    \item[] Question: Does the paper fully disclose all the information needed to reproduce the main experimental results of the paper to the extent that it affects the main claims and/or conclusions of the paper (regardless of whether the code and data are provided or not)?
    \item[] Answer: \answerYes{}
    \item[] Justification: The paper has disclosed all the necessary information to reproduce the main experimental results, including model architecture (Sec.~\ref{sec:method}), algorithm (Sec.~\ref{app:algorithm}), and implementation details (Sec.~\ref{app:implementation}).

\item {\bf Open access to data and code}
    \item[] Question: Does the paper provide open access to the data and code, with sufficient instructions to faithfully reproduce the main experimental results, as described in supplemental material?
    \item[] Answer: \answerYes{}
    \item[] Justification: We provide our code in the GitHub repository: \texttt{https://github.com/Skyyyyyalker/StructEvo}.

\item {\bf Experimental setting/details}
    \item[] Question: Does the paper specify all the training and test details (e.g., data splits, hyperparameters, how they were chosen, type of optimizer) necessary to understand the results?
    \item[] Answer: \answerYes{}
    \item[] Justification: The experimental settings are presented in Sec.~\ref{sec:exp-4site} and Sec.~\ref{sec:exp-full-len}.

\item {\bf Experiment statistical significance}
    \item[] Question: Does the paper report error bars suitably and correctly defined or other appropriate information about the statistical significance of the experiments?
    \item[] Answer: \answerYes{}
    \item[] Justification: We report standard deviation on results across 5 independent runs.

\item {\bf Experiments compute resources}
    \item[] Question: For each experiment, does the paper provide sufficient information on the computer resources (type of compute workers, memory, time of execution) needed to reproduce the experiments?
    \item[] Answer: \answerYes{}
    \item[] Justification: We provide the needed computer resources in Sec.~\ref{app:implementation}.
    
\item {\bf Code of ethics}
    \item[] Question: Does the research conducted in the paper conform, in every respect, with the NeurIPS Code of Ethics \url{https://neurips.cc/public/EthicsGuidelines}?
    \item[] Answer: \answerYes{}
    \item[] Justification: We conduct our work in accordance with the NeurIPS Code of Ethics.

\item {\bf Broader impacts}
    \item[] Question: Does the paper discuss both potential positive societal impacts and negative societal impacts of the work performed?
    \item[] Answer: \answerYes{}
    \item[] Justification: The societal impacts are in Sec.~\ref{sec:conclusion}.
    
\item {\bf Safeguards}
    \item[] Question: Does the paper describe safeguards that have been put in place for responsible release of data or models that have a high risk for misuse (e.g., pre-trained language models, image generators, or scraped datasets)?
    \item[] Answer: \answerNA{}
    \item[] Justification: We do not release a high-risk pretrained generative model; generated candidates require oracle evaluation, wet-lab validation, and human inspection before use.

\item {\bf Licenses for existing assets}
    \item[] Question: Are the creators or original owners of assets (e.g., code, data, models), used in the paper, properly credited and are the license and terms of use explicitly mentioned and properly respected?
    \item[] Answer: \answerYes{}
    \item[] Justification: We cite the original paper and clearly state assets details at Appendix~\ref{app:4site-benchmarks} and ~\ref{app:full-length-benchmarks}.

\item {\bf New assets}
    \item[] Question: Are new assets introduced in the paper well documented and is the documentation provided alongside the assets?
    \item[] Answer: \answerNA{}
    \item[] Justification: This paper does not release new assets.

\item {\bf Crowdsourcing and research with human subjects}
    \item[] Question: For crowdsourcing experiments and research with human subjects, does the paper include the full text of instructions given to participants and screenshots, if applicable, as well as details about compensation (if any)? 
    \item[] Answer: \answerNA{}
    \item[] Justification: This paper does not involve crowdsourcing nor research with human subjects.

\item {\bf Institutional review board (IRB) approvals or equivalent for research with human subjects}
    \item[] Question: Does the paper describe potential risks incurred by study participants, whether such risks were disclosed to the subjects, and whether Institutional Review Board (IRB) approvals (or an equivalent approval/review based on the requirements of your country or institution) were obtained?
    \item[] Answer: \answerNA{}
    \item[] Justification: This paper does not involve crowdsourcing nor research with human subjects.

\item {\bf Declaration of LLM usage}
    \item[] Question: Does the paper describe the usage of LLMs if it is an important, original, or non-standard component of the core methods in this research? Note that if the LLM is used only for writing, editing, or formatting purposes and does \emph{not} impact the core methodology, scientific rigor, or originality of the research, declaration is not required.
    \item[] Answer: \answerNA{}
    \item[] Justification: The core method development in this research does not involve LLMs as any important, original, or non-standard components.

\end{enumerate}

\end{document}